\documentclass{article}

\usepackage[eandd, final]{neurips_2026}

\usepackage[utf8]{inputenc} 
\usepackage[T1]{fontenc}    
\usepackage{hyperref}       
\usepackage{url}            
\usepackage{booktabs}       
\usepackage{amsfonts}       
\usepackage{nicefrac}       
\usepackage{microtype}      
\usepackage{xcolor}         
\usepackage{comment}
\usepackage{enumitem}
\usepackage{bm}
\usepackage{amsmath}
\usepackage{wrapfig}
\usepackage{graphicx}
\usepackage{multirow}

\title{Can We Trust Item Response Theory\\for AI Evaluation?}

\author{%
  Han Jiang$^{1}$ \quad Sunbeom Kwon$^{2}$ \quad Jinwen Luo$^{3}$ \quad Ziang Xiao$^{1}$ \quad Susu Zhang $^{2}$ \\
  $^1$Johns Hopkins University \quad $^2$University of Illinois Urbana-Champaign \\
  $^3$University of California, Los Angeles \\
  \texttt{hjiang66@jh.edu, szhan105@illinois.edu}}

\begin{document}

\maketitle

\begin{abstract}
  AI benchmarks increasingly leverage item-level statistical models, particularly item response theory (IRT), to estimate capabilities, rank systems, select informative examples, and diagnose benchmark quality. 
  However, AI benchmark data often departs from the data regimes encountered in human testing, for which standard IRT estimation tools were originally developed: these benchmarks typically involve fewer evaluated models, far more items, and capability distributions that may be skewed, clustered, or multimodal. 
  We examine how these regime mismatches challenge the reliability of IRT modeling for AI evaluation. 
  Using item parameters and capability distributions derived from six widely used LLM benchmarks, we simulate response matrices under three common IRT models and compare four estimators used in recent benchmark studies: marginal maximum likelihood, Markov chain Monte Carlo, variational inference, and a neural pseudo-Siamese estimator. 
  Across 18,000 simulation runs, we systematically evaluate computational feasibility, scalability, and the reliability of IRT inferences about model rankings, predicted performance, and item characteristics. 
  Results show that classical estimators can become computationally infeasible for large benchmarks, whereas scalable estimators can produce unreliable item-level and ranking inferences with small or non-normally distributed model sets. 
  This study identifies when latent trait models reliably support or risk distorting AI benchmarking claims, and what sample sizes and diagnostics are needed for trustworthy use.
\end{abstract}

\section{Introduction}\label{sec:1}

AI benchmarks have become the primary instrument for evaluating model capabilities, comparing AI systems, and informing deployment decisions.
Yet the dominant practice of relying on aggregate metrics such as mean accuracy is deceptively simple: 
it reduces a rich response matrix to a single number, treating all items as equally informative regardless of their difficulty or discriminative power~\citep{rodriguez-etal-2021-evaluation,ethayarajh-jurafsky-2020-utility}.
This simplification dilutes informative signals in evaluation data, renders model rankings sensitive to benchmark composition, and hinders the accumulation of comparable findings across evaluation efforts, reflecting a deep tension between validity, reliability, and efficiency~\citep{hofmann2025fluid,gill-etal-2025-lost}.

This tension naturally directs attention to psychometrics, the discipline devoted to the science of human testing and assessment, for well-established measurement tools. \textit{Item response theory} (\textbf{IRT}), the most widely adopted statistical framework, offers a principled alternative.
Instead of treating each item equally, IRT jointly estimates parameters of item characteristics and examinee ability on a shared latent scale, enabling in-depth diagnosis of item quality, efficient testing through adaptive item selection, and cross-benchmark comparison of ability estimates~\citep{lord1980applications}. 
Motivated by these advantages, a growing body of work has adopted IRT for AI evaluation, applying it to produce finer-grained model and item diagnostics (e.g.,~\cite{Kipnis2024-ex,schilling-wilhelmi2025lifting}) and enable more efficient evaluation strategies, such as adaptive AI testing (e.g.,~\cite{Jiang2025-jc,Truong2025-zj}) and benchmark compression (e.g.,~\cite{Polo2024-mu}).

\begin{wrapfigure}{r}{0.5\textwidth}
    \centering
    \includegraphics[width=\linewidth]{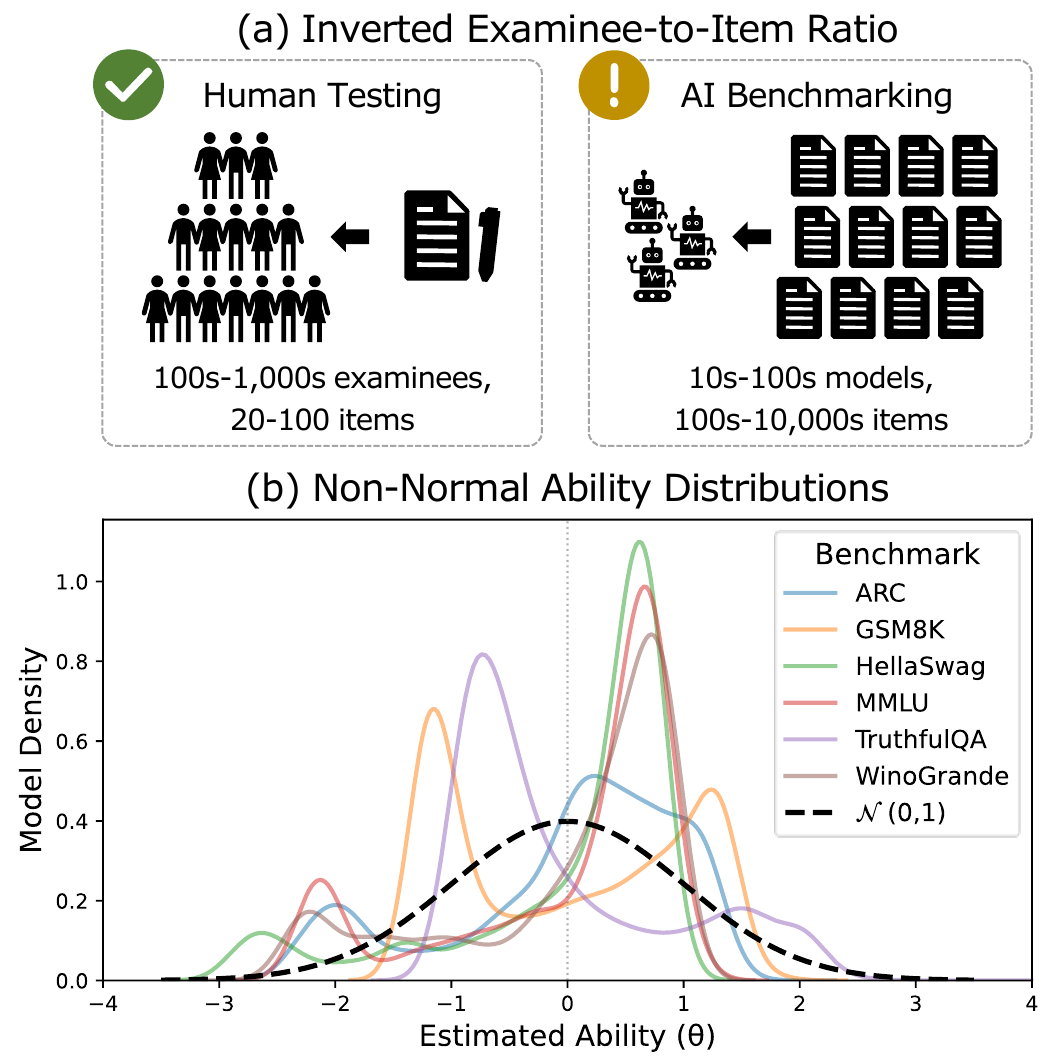}
    \caption{Illustration of the data regime mismatch between human testing and AI benchmarking. 
    }
    \label{fig:teaser}
\end{wrapfigure}

Nevertheless, the adoption has outpaced validation. 
A crucial concern is that \textit{AI benchmark data fundamentally depart from the human testing regime for which IRT estimation methods were developed and validated}. 
As illustrated in Fig.~\ref{fig:teaser}, the number of evaluated models is typically small (fewer than 100) while the number of items can reach tens of thousands. This differs from typical psychometric settings, where hundreds to thousands of examinees respond to tests of 20-100 items~\citep{de2013theory}.
Furthermore, the model capability distributions are frequently non-normal, exhibiting multimodality or skewness that conflicts with the normality assumptions underlying standard estimation procedures~\citep{lost,siska-etal-2024-examining}.
\textit{These departures have important implications,} as many IRT estimation methods' statistical validity directly depends on the sample-to-parameter ratio and the match between the assumed and the actual trait distribution. 
Violations of these assumptions may systematically bias inferences on both item characteristics and model capabilities.

Existing studies that apply IRT do not focus on verifying whether the chosen IRT models and estimation methods produce reliable estimates under this regime mismatch, leaving critical questions unanswered.
It remains unknown how small model populations, non-normal capability distributions, and long benchmarks affect parameter recovery and downstream inferences such as model rankings and item-level diagnostics.
Practical guidance for conducting such studies is equally lacking: for instance, when constructing an item-level benchmark repository (e.g., ~\cite{jiang2026position}), how many evaluated models will be sufficient to fit a given IRT model reliably?

To the best of our knowledge, no prior work has systematically addressed these questions. This paper fills this gap through three contributions:
\vspace{-2pt}
\begin{itemize}[itemsep=0.5pt, parsep=0pt, topsep=0pt,leftmargin=2em]
    \item A \textbf{\textit{literature review}} surveying 19 studies that apply IRT to AI benchmarking, characterizing the current landscape of IRT usage and empirically identifying the range of application conditions that inform the design of our simulation study (Sec.~\ref{sec:3}).
    \item A \textbf{\textit{simulation study}}\footnote{Code and data: \url{https://github.com/Salomeeeee/IRT-Simulation}.} calibrated to representative AI evaluation data regimes, systematically evaluating computational feasibility, parameter recovery, and downstream inference validity across six benchmarks, three IRT models, four estimators, and five sample sizes (Secs.~\ref{sec:4}~\&~\ref{sec:5}).
    \item \textbf{\textit{Practical guide}} based on the findings, including recommendations on minimum sample sizes, IRT model selection, and estimation methods, clarifying when IRT reliably supports AI benchmarking claims and when it risks distorting them (Sec.~\ref{sec:6}).
\end{itemize}
\vspace{-2pt}

Sec.~\ref{sec:2} provides the background on IRT and simulation methods, and Sec.~\ref{sec:7} gives concluding remarks.
\section{Background}\label{sec:2}

\subsection{Item response theory}\label{sec:2.1}

The limitations of aggregate metrics for AI evaluation naturally motivate the adoption of \textit{item response theory} (\textbf{IRT}), a family of statistical models from psychometrics. 
Unlike classical test theory which primarily relies on aggregate scores~\citep{hambleton1993comparison,gulliksen1950theory},  IRT explicitly accounts for differences in item difficulty by jointly modeling item characteristics and examinee ability on a common latent scale~\citep{lord1980applications, hambleton1991fundamentals}.

Specifically, IRT models the probability of the correct response to an item as a function of a latent ability parameter $\theta$ and one or more item-level parameters capturing item characteristics~\citep{lord1968statistical}. 
Its simplest form, the \textit{one-parameter logistic model} (\textbf{1PL}, also known as the \textit{Rasch} model~\citep{rasch1960probabilistic}), characterizes each item solely by its \textit{difficulty} $b$: $p(y\!=\!1|\theta, b)=\sigma(\theta\!-\!b)$, where $y\!\in\!\{0, 1\}$ denotes the response correctness and $\sigma(\cdot)$ is the sigmoid function.
The \textit{two-parameter logistic model} \textbf{(2PL}) introduces a \textit{discrimination} parameter $a$, allowing items to differ in how effectively they distinguish between high- and low-ability examinees: $p(y\!=\!1|\theta, a,b)=\sigma(a(\theta\!-\!b))$.
The \textit{three-parameter logistic model} (\textbf{3PL}) further adds a \textit{pseudo-guessing} parameter $c$, representing the probability that an extremely low-ability examinee responds correctly: $p(y\!=\!1|\theta, a,b)=c\!+\!(1\!-\!c)\sigma(a(\theta\!-\!b))$~\citep{Lord1966SomeLT}.
Extensions such as the 4PL~\citep{4pl} and the graded response model~\citep{samejima1969estimation} incorporate additional parameters into the response function.

In practice, IRT models are estimated using observed response data from settings such as educational testing and AI benchmarking. 
Given a response matrix $Y \in \{0,1\}^{N\!\times\!J}$ where each entry $y_{i,j}$ records whether examinee $i$ answered item $j$ correctly, an IRT model is fit to jointly estimate $\theta_i$ for each examinee and the item parameter(s) $\bm \zeta_j=\{b_j, ...\}$, for each item.
Standard estimators include marginal maximum likelihood via Expectation-Maximization~(\textbf{MML-EM}~\citep{bock1981marginal}), Markov chain Monte Carlo~(\textbf{MCMC}~\citep{mcmc1,mcmc2}), and variational inference~(\textbf{VI}~\citep{wu2020variationalitemresponsetheory}), which we elaborate on in Sec.~\ref{sec:3}.
Estimated parameters inform downstream analyses, including flagging problematic items, constructing shorter yet equally informative tests, and ranking examinees on a common scale with associated uncertainty.

Despite these benefits, we highlight two characteristics of AI benchmarking data that challenge the assumptions of common IRT estimation methods, which \textit{potentially compromise the reliability of IRT-based AI benchmark analysis}.
First, standard IRT estimators were developed primarily for settings with relatively large examinee samples and comparatively few items. Accurate estimation typically requires at least $N = 500$ for 2PL and $N=1,000$ for 3PL~\citep{de2013theory}. Statistical consistency of MML-EM further assumes fixed $J$ and $N\to\infty.$ In contrast, AI benchmarks often evaluate fewer than 100 models on up to tens of thousands of items~\citep{pmlr-v238-frick24a}.

Second, standard estimators commonly assume normally distributed abilities, either explicitly (MML-EM) or through prior specification (Bayesian and VI).
However, model capability distributions may exhibit substantial skewness or multimodality, bearing little resemblance to the assumed normal distributions~\citep{lost,siska-etal-2024-examining}. Evaluated models rarely constitute a random sample
from a well-defined population. Instead, AI benchmarks typically comprise community-contributed convenience samples, in which many models share foundation models and adaptation pipelines \citep{bommasani2022opportunities,NEURIPS2023_a0b1082f}, forming a non-random, lineage-dependent convenience set with uneven representation across architecture families, training regimes, and capability strata.

\subsection{Simulation studies: evaluating the evaluator}

Simulation studies, also known as Monte Carlo studies, are the standard methodology for evaluating the performance of statistical methods under controlled conditions~\citep{morris2019using}. 
The general procedure follows a well-established workflow: 
\vspace{-2pt}
\begin{itemize}[itemsep=0.5pt, parsep=0pt, topsep=0pt,leftmargin=2em]
    \item specify true values for parameters that are ordinarily unobservable, 
    \item generate simulated response data from these known parameters and model, 
    \item fit the model of interest to the simulated data, and 
    \item compare the estimated parameters against the data-generating ground truth,
\end{itemize}
\vspace{-2pt}
which has been formalized as the ADEMP framework -- \underline{A}ims of the study, \underline{D}ata generation mechanisms, \underline{E}stimated parameters, \underline{M}ethods being evaluated, and \underline{P}erformance measures -- for planning and reporting simulation studies~\citep{siepe2024simulation}. 
By directly manipulating simulation conditions and evaluating estimation accuracy against known ground truth, we can isolate the effects of specific conditions on method performance.
An IRT simulation study thus serves as a means of evaluating the evaluator itself: by systematically varying simulation conditions such as the number of evaluated models, the distribution of model capabilities, the size of AI benchmarks, and the choice of IRT models, we can directly examine whether the resulting estimates (i.e., model capabilities and item parameters) reliably support their intended uses.

In psychometrics, a substantial body of simulation research has investigated the conditions under which IRT estimates for human testing are trustworthy (e.g.,~\cite{10.3389/fpsyg.2016.00109,Sen2023-ye}), establishing the practical guidelines and assumptions outlined in Sec.~\ref{sec:2.1}.
For example, recommended minimum sample sizes are 250-500 for 1PL and 2PL~\citep{de2013theory,demars2010item,schroeders2025sample}, and 1,000-3,000 for 3PL~\citep{de2013theory}.
These studies also document how assumption violations, such as non-normal ability distributions or short test lengths, degrade estimation quality (e.g.~\cite{stone1992recovery,woods2006item,kirisci2001robustness}). 
However, these guidelines were developed under the human testing conditions described above. 
AI evaluation setting departs from these conditions, yet no prior work has conducted simulation studies in this context. 
This paper thus addresses this gap with a comprehensive simulation study calibrated to representative AI evaluation conditions observed in our literature review.
\section{Item response theory for AI evaluation}\label{sec:3}

Taking the simple average score on a benchmark ignores that not all benchmark items are equally informative~\cite{rodriguez-etal-2021-evaluation}. 
Items with poor discriminative power, such as those on which models perform similarly regardless of overall capability, do not contribute capability-relevant signal~\cite{heineman2026signal}. 
As models rapidly evolve and more items become trivially easy, this becomes increasingly common~\cite{ott2022benchmark,deveci2025the}. 
Additionally, some items may capture shortcuts such as spurious correlations or format heuristics rather than the intended capability, jeopardizing the validity of score interpretations.
Without principled methods to identify and down-weight such uninformative items, they contribute equally to aggregate scores alongside more diagnostic items. 
Furthermore, benchmark-specific average score lacks mechanisms to place models evaluated on different benchmarks on a common scale~\cite{liang2023holistic}, hindering the accumulation of comparable findings across evaluation efforts.

IRT emerged as a viable approach to addressing these AI benchmarking challenges. As a survey of current IRT applications to address these limitations, we conducted a scoping review.
We searched arXiv using keywords related to \textit{item response theory (IRT)}, \textit{AI evaluation}, and \textit{AI benchmarking}, covering publications from 2016 to April, 2026.
Two authors independently screened the results against predefined inclusion criteria: studies must apply IRT or related item-factor models \textbf{to} model evaluation in AI, ML, or NLP contexts; studies that merely cite IRT without applying it were excluded.
This process identified 19 studies.

Across these studies, IRT-style probablistic models have been used to support 
(1) \textit{latent capability scoring} that accounts for item difficulty or discrimination in place of simple aggregate scores \cite{lalor-etal-2016-building, Martinez-Plumed2019-jv, rodriguez-etal-2021-evaluation, Kipnis2024-ex}; 
(2) \textit{item and benchmark analysis}, including the identification of difficult, informative, saturated or weakly discriminating examples \cite{Lalor2018-gn,Lalor2019-uw, rodriguez-etal-2021-evaluation,Byrd2022-cq, schilling-wilhelmi2025lifting, lost}; and 
(3) \textit{efficient evaluation} through smaller item subsets, adaptive item administration or generation, and prediction of performance on unseen items or tasks, reducing the need to run every model on full benchmarks \cite{Zhuang2023-kk, Polo2024-mu, Kipnis2024-ex,Truong2025-zj,hofmann2025fluid,Jiang2025-jc,Liao2025-wn, Chen2025-aa}. 
IRT model specifications ranged from standard binary 1PL, 2PL, 3PL and 4PL, with either unidimensional or multidimensional latent ability spaces, to adaptations for data formats including pairwise comparisons (e.g., \cite{Robertson2025-ld}), continuous scores (e.g., \cite{Liao2025-wn}), response-length information \cite{Xu2025-ye}, and item text features (e.g., \cite{Chen2025-aa}).
The \textit{unidimensional 2PL with item-specific difficulty and discrimination parameters} was the most commonly adopted model. 

Most of these studies fitted IRT models to binary response matrices, in which rows represent evaluated models and columns represent benchmark examples, or items. However, the row and column dimensions differed substantially from typical human-testing contexts and varied substantially across studies. 
Many included tens to low hundreds of evaluated models, including cases below 50, with analyses based on OpenLLM Leaderboard~\cite{myrzakhan2024openllm} as the main exception, where model counts exceeded 5,000. 
Benchmark length, by contrast, ranged from a few hundred to more than 80,000 examples. 
Estimation methods were also heterogeneous, including marginal or full-information maximum likelihood via the EM algorithm \cite{Truong2025-zj, Kipnis2024-ex}, MCMC \cite{hofmann2025fluid, Liao2025-wn, schilling-wilhelmi2025lifting}, variational inference \cite{vania-etal-2021-comparing, rodriguez-etal-2021-evaluation, Polo2024-mu, Jiang2025-jc}, and neural-network representations of IRT models fit with gradient-based optimization \cite{lost,Jiang2025-jc}. 

This variation motivates a methodological study that examines how evaluated model count, benchmark length, capability distribution, IRT model specification and estimation method affect the reliability of item- and model-level inferences used to construct and interpret AI benchmark claims.
\section{Simulation design}\label{sec:4}
The simulation aims to examine \textit{whether existing IRT estimation approaches provide reliable item- and model-level inferences under data regimes observed in AI evaluation}.
The data conditions, estimation methods, and evaluation metrics were chosen to reflect common practices in recent applications of IRT to AI benchmark analysis. 

\subsection{Data generation}\label{sec:4.1}

\paragraph{IRT model, sample size, and benchmark.} 
For IRT model, we focus on unidimensional 1PL, 2PL, and 3PL as the most commonly adopted specifications in the reviewed studies. 

Sample sizes, i.e., the number of evaluated models, were chosen from the empirical range observed across the studies in Sec.~\ref{sec:3}:
$N\!=\!30$, similar to sample sizes found in \cite{lost, Polo2024-mu, Truong2025-zj}; $N\!=\!100$, similar to \cite{Jiang2025-jc, vania-etal-2021-comparing, Polo2024-mu, hofmann2025fluid}; $N\!=\!180$, similar to \cite{rodriguez-etal-2021-evaluation, Truong2025-zj}; $N\!=\!400$, similar to \cite{Polo2024-mu}; and $N\!=\!1000$, as a reference condition closer to sample-size recommendations for 3PL estimation in human testing.

Benchmark conditions were based on the OpenLLM Leaderboard v1 benchmarks analyzed in \cite{Polo2024-mu, hofmann2025fluid, Kipnis2024-ex}. 
The leaderboard aggregates results from over 6,000 community-contributed models across six benchmark datasets: \textsc{ARC-Challenge}~\cite{clark2018arc}, \textsc{HellaSwag}~\cite{zellers-etal-2019-hellaswag}, \textsc{MMLU}~\cite{hendryckstest2021}, \textsc{TruthfulQA}~\cite{lin-etal-2022-truthfulqa}, \textsc{WinoGrande}~\cite{10.1145/3474381}, and \textsc{GSM8K}~\cite{cobbe2021gsm8k}, spanning tasks from commonsense reasoning to mathematical problem solving, with benchmark lengths ranging from approximately 800 to 15,000 items. 
In total, this yields $3\!\times\!5\!\times\!6\!=\!\bm{90}$ unique simulation conditions.

\paragraph{True parameters and response matrices.} 
The true item parameters and model capability distributions were derived from the preprocessed OpenLLM response matrices used in \cite{Kipnis2024-ex}, which contain binary responses from over 6,000 models across the six benchmarks. 
Following their preprocessing, models with the lowest 0.1\% of scores were removed to reduce noise, and items with standard deviations below 1\% or mean accuracies above 95\% were excluded as uninformative or nearly saturated. 
The retained matrices' dimensions ($N$ models $\times J$ items) were as follows: $5,222\!\times\!644$ for \textsc{TruthfulQA}, $5,221\!\times\!844$ for \textsc{ARC-Challenge}, $6,550\!\times\!1,045$ for \textsc{WinoGrande}, $6,068\!\times\!1,306$ for \textsc{GSM8K}, $5,221\!\times\! 5,711$ for \textsc{HellaSwag}, and $5,219\!\times\!12,508$ for \textsc{MMLU}. 

For each benchmark, we fit the 1PL, 2PL and 3PL models to the empirical response matrix using variational inference (VI), given its fast computation in high-$N$, high-$J$ settings.
The resulting item and model capability parameters were first rescaled so that the capability distribution has mean 0 and variance of 1. Then, the item parameter estimates were treated as the \textit{\textbf{true item parameters}} in the simulation. 
Figs.~\ref{fig:app_tb_truth}--\ref{fig:app_tb_mmlu} in App.~\ref{app:theta_beta} present the distributions of the estimated model capabilities and true difficulty parameters for each benchmark, illustrating the preserved distributional features such as skewness, clustering, and multimodality.
Additionally, to assess whether using the VI-estimated parameters as generating values favors VI in estimator comparisons, we conducted additional simulations with PSN-IRT-estimated parameters. 
The results in App.~\ref{app:true_params} showed similar relative estimator performance, except for item difficulty recovery.

For each of the 90 conditions, we generated 50 replications. 
In each replication,  $N$ \textit{\textbf{true capability values}} were sampled from the empirical distribution of the estimated capabilities for the corresponding condition. 
The sampled capability values and the item parameters were then used to generate a full binary response matrix, which was used as the input data for IRT estimation.

\subsection{Estimation}\label{sec:4.2}
We evaluated four IRT estimators spanning the major paradigms used in recent IRT-based AI evaluation studies (Sec.~\ref{sec:3}): MML-EM, MCMC, VI, and PSN. Full technical details, including formulations, prior specifications, and implementation settings, are provided in App.~\ref{app:estimation}.

\textbf{MML-EM} estimates item parameters by maximizing the marginal likelihood, integrating out latent abilities under an assumed population distribution, typically the normal distribution. 
It is the conventional approach in psychometrics, implemented here via the Metropolis-Hastings Robbins-Monro algorithm, a stochastic variant of EM, in the \texttt{mirt} R package~\citep{mirt}. 
MML assumes a fixed, known ability distribution and its statistical consistency guarantees require $N \to \infty$ with $J$ fixed, which is inverted in AI benchmarking.
\textbf{MCMC} performs full Bayesian inference, sampling from the joint posterior over item parameters and examinee abilities. We used Stan's Hamiltonian Monte Carlo via the \texttt{brms} R package~\citep{brms}. MCMC iteratively samples parameters from the posterior distribution with computational demand scaling in $N$, $J$ and chain length. 
\textbf{VI} approximates the Bayesian posterior with a factorized variational distribution, optimized via KL divergence minimization. 
We used the implementation in the Python package, \texttt{py-irt}~\citep{pyirt}. 
VI is computationally attractive for larger response matrices as it replaces the posterior sampling with gradient-based optimization on the approximate distribution. 
However, it inherits the prior assumptions of the Bayesian model and adds approximation error from the chosen variational family.
\textbf{PSN} is a neural-network estimator that learns ability and item parameter embeddings, which are passed through an IRT response function to predict binary responses, trained with gradient-based optimization~\citep{lost}. 
Unlike the other estimators, PSN does not impose a parametric distribution on abilities, but it does not directly provide uncertainty quantification or consistency guarantees for its estimates.

\subsection{Evaluation metrics}\label{sec:4.3}
The evaluation covers computational feasibility and the three IRT use cases identified in Sec.~\ref{sec:3}: latent capability scoring, item analysis, and efficient benchmarking, resulting in four corresponding subsections in Sec.~\ref{sec:5}.

\paragraph{Computational feasibility.} Computational feasibility was summarized by \textit{\textbf{failure rate}} (the proportion of replications that failed to produce parameter estimates) and per-replication \textit{\textbf{runtime}} for completed runs. 
For MML-EM, failures also included non-convergence when the algorithm terminated without satisfying the convergence criterion after 2,000 cycles.

\paragraph{Latent capability scoring.} For model capability, \textbf{\textit{ranking recovery}} was evaluated using Kendall's rank ($\tau$) correlation between true and estimated capability values. 
Additionally, we measured \textbf{\textit{aggregate score error}}: the absolute difference between expected full-benchmark scores under true and estimated parameters, $\left|\frac{1}{J}\sum_{j=1}^{J}\pi_{ij}-\frac{1}{J}\sum_{j=1}^{J}\hat{\pi}_{ij}\right|$, averaged across models, where $\pi_{ij}=P(Y_{ij}=1\mid \theta_i,\boldsymbol{\zeta}_j)$ and $\hat{\pi}_{ij}=P(Y_{ij}=1\mid \hat{\theta}_i,\hat{\boldsymbol{\zeta}}_j)$ denote model $i$'s correct response probability on item $j$ based on true and estimated parameters, respectively. 

\paragraph{Item analysis.} Recovery of item-level properties was evaluated using Pearson correlations between true and estimated item parameters, including \textbf{\textit{difficulty recovery}} and \textbf{\textit{discrimination recovery}} (2/3PL), and the average absolute error in estimated guessing probability (3PL), i.e., $\frac{1}{J}\sum_{j=1}^{J}|\hat c_j-c_j|$. 
Similar to aggregate score error, we measured \textbf{\textit{item mean error}} as absolute deviation between item-level correct response probabilities computed from estimated and true parameters, $\left|\frac{1}{N}\sum_{i=1}^{N}\pi_{ij}-\frac{1}{N}\sum_{i=1}^{N}\hat{\pi}_{ij}\right|$, averaged across items.

\paragraph{Efficient benchmarking.} To assess IRT's effectiveness in AI benchmark compression, Fisher information $\mathcal{F}$~\cite{lord1980applications} was computed for each item $j$ given each model's estimated capability value $\hat{\theta}_i$:
$
\hat{\mathcal{F}}_{j}(\hat{\theta}_i)
=
\hat a_j^2 \cdot
\frac{(\hat{\pi}_{ij}-\hat c_j)^2}{(1-\hat c_j)^2} \cdot
\frac{1-\hat{\pi}_{ij}}{\hat{\pi}_{ij}} .
$
For 1PL and 2PL, $\hat c_j=0$, and for 1PL, $\hat a_j= 1$. 
The top 10\%, 20\%, and 100 items with largest Fisher information averaged across $\hat{\theta}_i$ formed the short benchmark. 
Using 4,000 held-out models not used for IRT estimation, \textbf{\textit{short-form ranking recovery}} was computed as Kendall correlation between average scores on the short benchmark and the full benchmark. 
This measures whether IRT-based item selection can support benchmark compression while preserving full-benchmark model rankings.
\section{Results}\label{sec:5}
\subsection{Which IRT estimators are computationally feasible?}\label{sec:5.1}
\begin{figure*}
    \centering
    \includegraphics[width=0.9\linewidth]{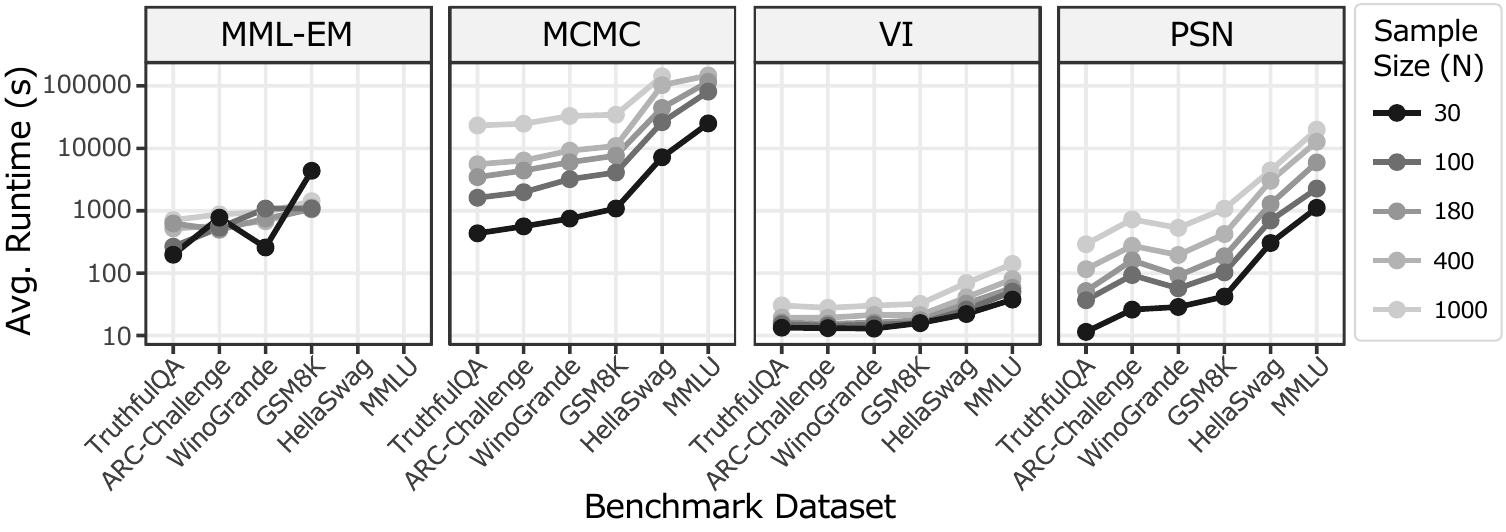}
    \caption{Per-replication runtime (in seconds, log scale) averaged over 1PL, 2PL, \& 3PL, across IRT estimators, benchmarks, and sample sizes. 
    Missing points for MML-EM and MCMC on \textsc{HellaSwag} ($J$=5,711) and \textsc{MMLU} ($J$=12,508) indicate computational infeasibility.}
    \label{fig:comp_time}
\end{figure*}

Across all conditions and replications, MML-EM exhibited the highest failure rate (69.45\%).
Specifically, for \textsc{HellaSwag} and \textsc{MMLU} with $J>$ 5,000, MML-EM consistently ran out of memory on 32GB CPU RAM. The out-of-memory error persisted on one tested replication with 128GB CPU RAM with \textsc{HellaSwag}, 1PL, $N = 30$ , rendering it infeasible for large-scale benchmarks. 3PL conditions consistently showed estimation or convergence failures. 
For smaller benchmarks under 1PL and 2PL, MML-EM often ran into convergence failures at small sample sizes.
The distributions of MML-EM and MCMC exit status are detailed in App.~\ref{app:exit}.

For MCMC, estimation became computationally prohibitive for the larger sample sizes and benchmarks: With a 72-hour limit, no replications finished for \textsc{MMLU} when $N = 1000$ (under any IRT model) and $N = 400$ (3PL), and for \textsc{HellaSwag} when $N=1000$ (3PL). $5.8\%$ of replications under other \textsc{MMLU} and \textsc{HellaSwag} conditions also hit the 72-hour limit. Less than $0.1\%$ of the finished replications within 72 hours failed due to algorithm divergence.

VI had an overall failure rate of 10.71\%. All failures were estimation algorithm failures, primarily in the 3PL conditions (27.56\% of replications), with slightly higher rates for larger benchmarks.  PSN achieved 0\% failure across all conditions. 

Runtime differed substantially across estimators (Fig.~\ref{fig:comp_time}). 
As 1PL, 2PL, and 3PL did not differ meaningfully in runtime within each estimator, we report averages across IRT models and replications. 
Note that these runtimes reflect practical feasibility under the hardware configurations tested rather than a controlled comparison of computational efficiency: MML-EM, MCMC, and VI were run on CPU, while PSN was run primarily on a single A100 GPU, with a small number of replications on H100 and V100.
VI was the fastest estimator with minimal sensitivity to sample size and benchmark length;
MCMC was the slowest, often exceeding 10,000 seconds per replication; and PSN runtime scaled with both benchmark length and sample size, growing steeply for large benchmarks; while manageable in our conditions, its scalability to very large benchmarks and samples warrants caution.

\subsection{How do different IRT estimators recover model capabilities and rankings?}\label{sec:5.2}
Ranking recovery (Kendall's $\tau$) fell below $.60$ in some conditions, with variation depending primarily on the specific benchmark–IRT model combination. Within a benchmark-IRT model combination, there was comparatively little effect of sample sizes or estimator.
Recall that each benchmark-IRT model combination shared the same true item parameters and capability distribution. To understand what specific item or capability distribution characteristics predicted poor ranking recovery, we fitted a multilevel regression model predicting Fisher-transformed Kendall's $\tau$ between true and estimated capabilities. The regression model included four continuous predictors: (i) absolute sample skewness statistic of the true capabilities \citep{skewness}; (ii) insufficient difficulty coverage, defined as the proportion of evaluated models with $5$ or fewer items whose difficulties were within \(0.2\) of their true capabilities; (iii) marginal reliability, where a lower value implies limited item discrimination, jointly reflecting difficulty mismatch and low item discrimination; and (iv) sample size, $\log(N)$. Continuous predictors were standardized, and estimator was included as a categorical predictor. Detailed definitions, model specifications, and results are provided in App.~\ref{app:failure_analysis}; the full model had $R^2=.798$.

\begin{figure*}
    \centering
    \includegraphics[width=\linewidth]{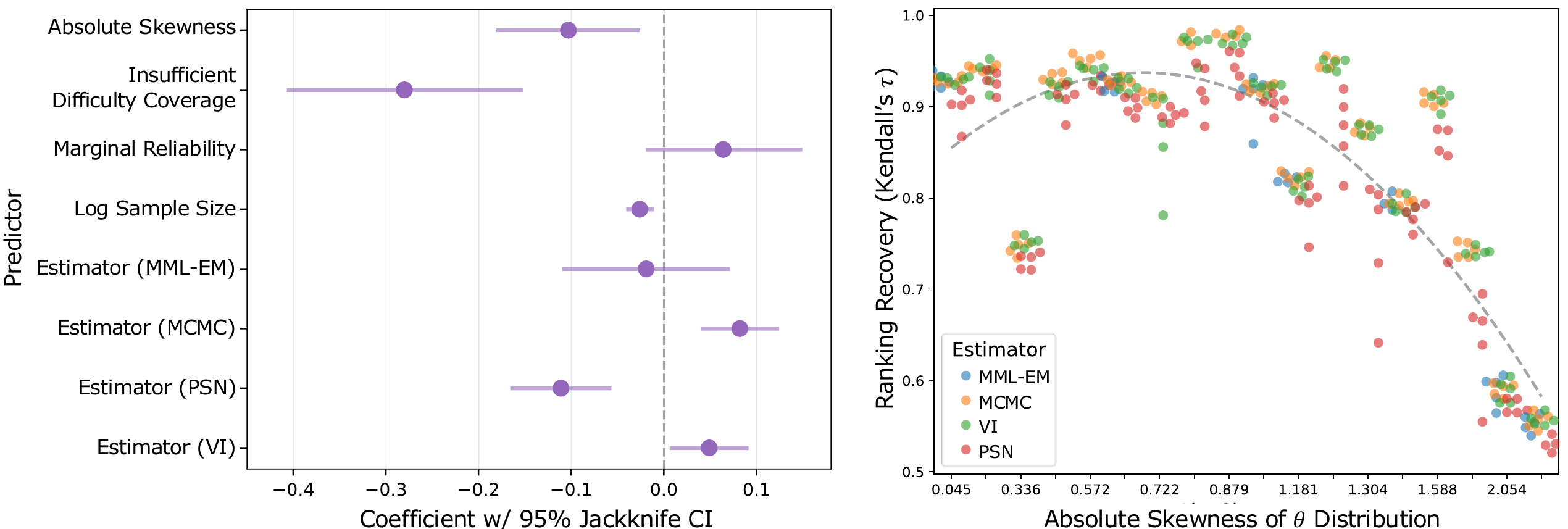}
    \caption{
    Factors (predictors) associated with ranking recovery. Left: Regression coefficients with 95\% jackknife CIs with intercept omitted. Right: Ranking recovery versus absolute capability skewness; points represent simulation conditions, colored by estimator, with sample sizes not visually distinguished. 
    Dashed lines indicate zero (left) and a quadratic fit (right).
    }
    \label{fig:skewness_ranking}
\end{figure*}

As shown in Fig.~\ref{fig:skewness_ranking} (left), \textit{insufficient difficulty coverage was the strongest predictor of poor ranking recovery} (coefficient $=-.280$). Adding it to a base model with only estimator and log sample size as covariates increased marginal $R^2$ from $.039$ to $.760$. \textit{Absolute capability skewness was also strongly predictive when added separately to the base model}, increasing marginal $R^2$ to $.437$. Capability skewness was moderately correlated with insufficient difficulty coverage (Kendall’s $\tau=.40$) and retained incremental predictive value for poor ranking (coefficient  = $-.103$) after accounting for insufficient coverage and the other predictors. Fig.~\ref{fig:skewness_ranking} (right) also shows how ranking recovery declines as the true capability distribution becomes more skewed. Marginal reliability had a small positive association with ranking recovery but had limited incremental predictive value. 

Comparing estimators, MCMC and VI performed above the estimator mean, whereas PSN performed below it. However, the effects of sample size and estimator were both relatively small, suggesting that when too few items target the difficulty range where the evaluated models differ in capability, or when true capability distribution is irregular, \textit{increasing sample size or choosing a particular estimator does not substantially improve ranking.}

Poor recovery of rankings does not necessarily imply practically consequential errors in estimated performance. Exact rankings may be less meaningful when models have nearly identical capabilities. In these settings, small estimation errors can reverse model ordering without materially affecting estimated performance. Consistent with tinyBenchmarks~\citep{Polo2024-mu}, aggregate score error (Fig.~\ref{fig:app_agg_err}) remained below $.025$ across most conditions, including skewed conditions such as \textsc{HellaSwag} and \textsc{TruthfulQA} where ranking recovery degraded sharply.

Nevertheless, rankings matter when benchmark results support comparative decisions, such as selecting a model for deployment \citep{jung2026defines} or determining whether a new model improves on a baseline \citep{heineman2026signal}. This is particularly relevant for general-purpose benchmarks whose results are interpreted as evidence of broader capabilities and guide downstream model selection \citep{salaudeen2025measurement}. Moreover, a commonly stated benefit of IRT in the studies reviewed in Sec.~\ref{sec:3} is that accounting for item difficulty and discrimination can better distinguish models with similar aggregate scores. Our ranking analysis therefore cautions against capability comparisons from the ordering of point estimates alone, especially when items provide limited information for distinguishing models and capability distribution is skewed.

\subsection{How do different IRT estimators recover item and benchmark characteristics?}\label{sec:5.3}

\begin{figure*}
    \centering    
    \includegraphics[width=0.9\linewidth]{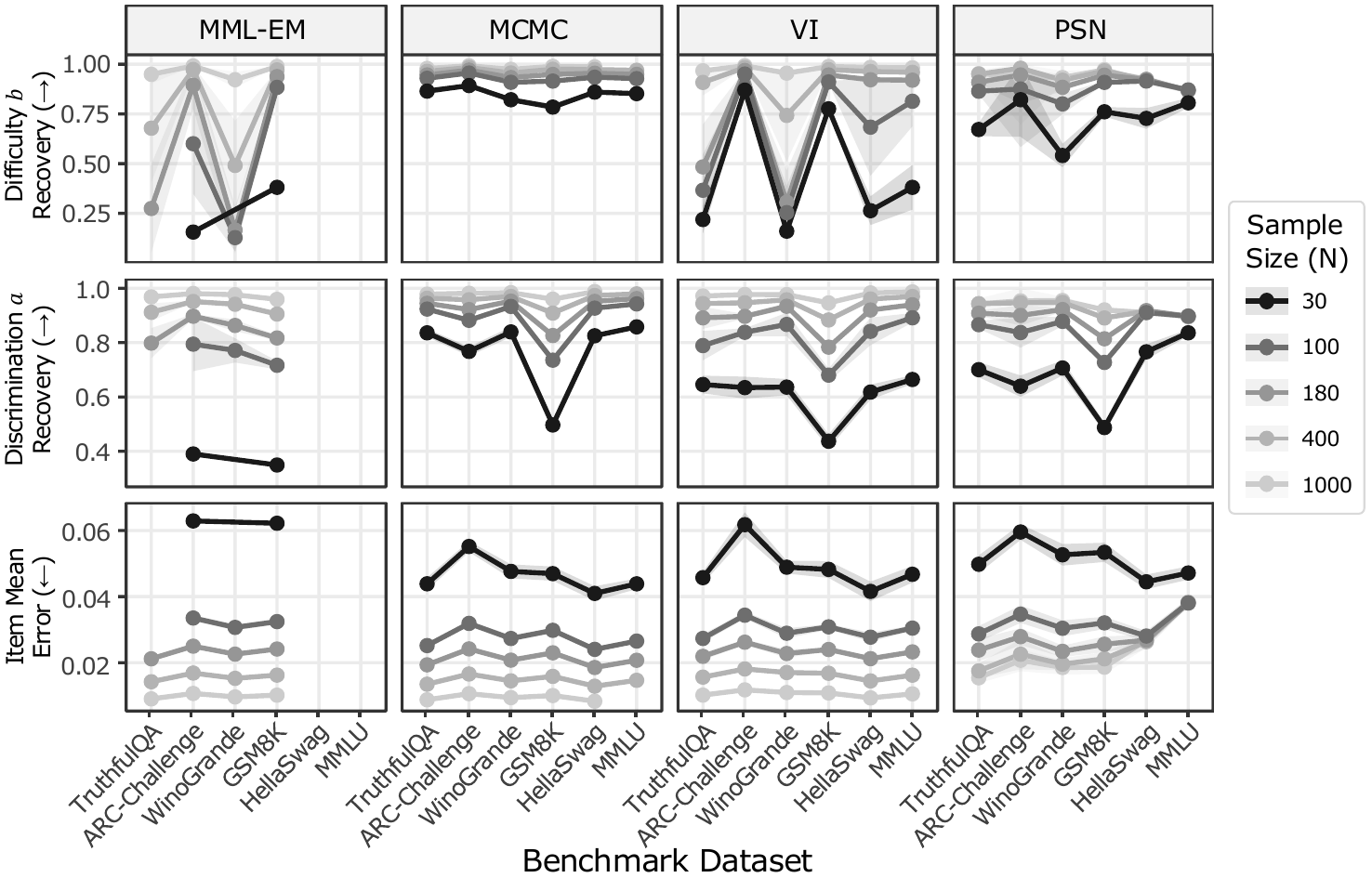}
    \caption{Item parameter recovery under 2PL across estimators, benchmarks, and sample sizes. 
    Higher is better for $b$ and $a$ recovery; lower is better for item mean error. 
    Shaded regions indicate $\pm$1 standard deviation across 50 replications. 
    Missing points indicate computational infeasibility.}
    \label{fig:item_recovery}
\end{figure*}

Here, we present results under 2PL as the commonly adopted IRT specification in AI benchmarking literature. Fig.~\ref{fig:item_recovery} presents the difficulty recovery, discrimination recovery, and item mean error across estimators, benchmarks, and sample sizes.
Full results for 1PL and 3PL are reported in App.~\ref{app:item_recovery}.

\textit{Reliable estimation of item parameters depended heavily on sample size}, i.e., the number of evaluated models: $N$ = 30 (the darkest lines) was insufficient for item parameter recovery under any estimator. 
At this sample size, difficulty recovery dropped below 0.50 (e.g., MML-EM on \textsc{ARC-Challenge} and PSN on \textsc{WinoGrande}) and discrimination recovery below 0.60 (e.g., MML-EM and PSN on \textsc{GSM8K}). Item mean error was also substantially higher compared to all larger sample sizes. 
A marked improvement occurred at $N$ = 100, though further increasing sample size led to incremental gains. This suggests that the small model counts ($N\approx 30$) commonly observed in AI benchmarking (Sec.~\ref{sec:3}) could risk item-level inference reliability.

A remaining question is whether $N=30$ would suffice for reliable item parameter recovery under more regular capability distributions. 
We conducted two supplementary analyses: 
(i) a comparison of $N=30$ replications with near-normal empirical distributions of true capabilities against their $N=100$ counterparts, and (ii) controlled 3PL simulations on \textsc{ARC-Challenge}, using capabilities drawn from a standard normal distribution or resampled from the empirical capabilities within $[-1,1]$.
The parameter recovery gap between $N$ = 30 and $N$ = 100 persisted in both analyses. Even under normal capability distributions, the examined estimators showed poor item-parameter recovery at $N=30$, with clear improvements at $N=100$.
More details are provided in App.~\ref{app:disentangle}.

Among the IRT estimators, MCMC generally achieved the most reliable item parameter recovery when computationally feasible.
VI, despite being one of the most widely used estimators in recent IRT-based AI evaluation studies, showed notably unreliable difficulty recovery in several conditions: For 2PL, its difficulty recovery was below $.50$ for $N\leq 180$ on \textsc{TruthfulQA} and \textsc{GSM8K}, indicating that VI-estimated item difficulties may diverge substantially from true item difficulty.
Similarly, MML-EM, the conventional approach to IRT estimation in human testing, produced unreliable estimates for small samples and certain benchmarks, even when it converged.
PSN, in general, showed more reliable item parameter recovery than MML-EM and VI, with slightly lower accuracy and higher variability across replications than MCMC in some conditions. 


\subsection{Can estimated IRT parameters support efficient benchmarking?}\label{sec:5.4}

\begin{wrapfigure}{r}{0.45\textwidth}
    \vspace{-12pt}
    \centering
    \includegraphics[width=\linewidth]{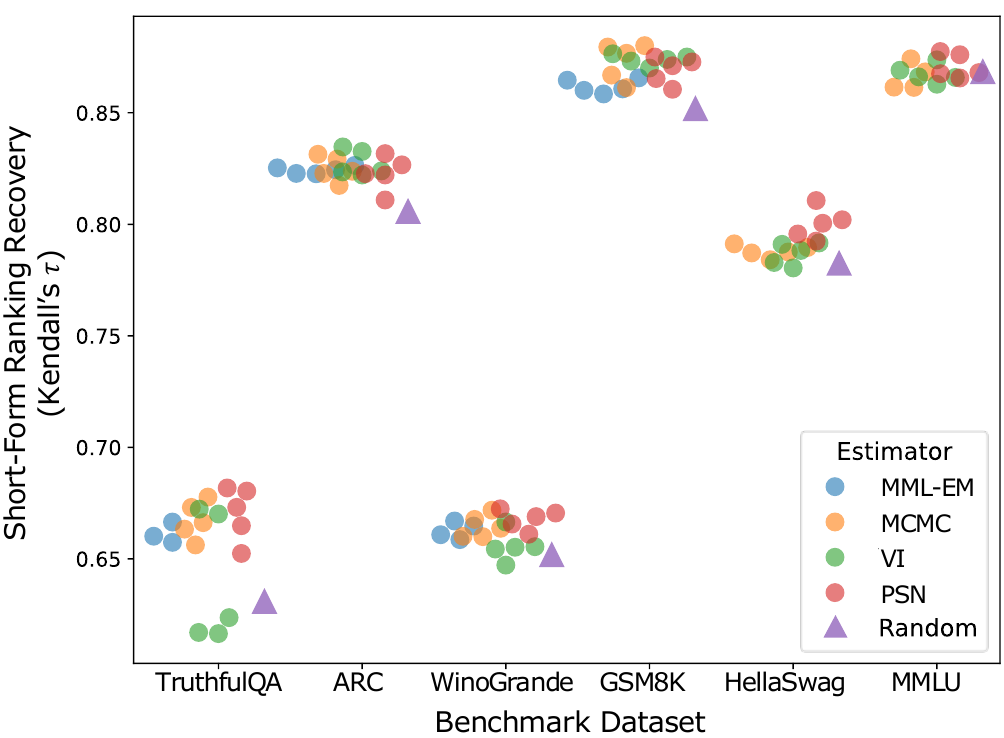}
    \caption{Short-form ranking recovery across benchmarks and estimators (2PL). Each point represents one simulation condition.
    }
    \label{fig:shortform}
\end{wrapfigure}

Finally, we examined whether short benchmarks, constructed based on expected Fisher information using estimated item parameters, could accurately predict aggregate score on the full test for held-out samples.
We evaluated subsets containing $10\%$ or $20\%$ of the full benchmark items, or a fixed length of $100$ items. Each was compared with a random-selection baseline averaged over 30 item subsets of the same size.
Across these subset sizes, IRT-based short benchmark generally outperformed the random baseline across benchmarks and estimators, with pronounced improvements under 1PL but only marginal improvements under 2PL and 3PL (Fig.~\ref{fig:app_efficient_grid}). 
For 1PL, larger samples ($N\geq100$) improved short-form ranking recovery, as shown in Figs.~\ref{fig:app_20p} \&~\ref{fig:app_100}.

Fig.~\ref{fig:shortform} shows the ranking recovery under 2PL for the $10\%$ item subsets. 
Unlike full-benchmark ranking recovery (Fig.~\ref{fig:skewness_ranking}), short-form ranking recovery showed no monotonic relationship with the skewness of capability distribution. 
To assess whether the modest gains under 2PL were due to the small subset sizes, we also examined larger subsets across all six benchmarks, including subsets of up to 2,500 items for MMLU. 
IRT-based short benchmarks continued to perform similarly to the random baselines.
Detailed results for all short benchmarks are provided in App.~\ref{app:short_forms}.

\section{Practical recommendations and implications}\label{sec:6}

Our results suggest that IRT-based AI benchmark analysis requires different safeguards depending on the intended use of the fitted model. 
For estimator choice, MML-EM is the conventional approach in human testing, but in our simulations it became computationally burdensome for large-item benchmarks and did not always yield reliable estimates even when runs converged. 
MCMC provided the most consistent parameter recovery, but its computational cost limited its feasibility for combinations of large benchmarks, large sample sizes, and more complex IRT models. 
VI was fast and recovered model rankings slightly better, but was found unreliable sometimes for aggregate score recovery and item-parameter recovery. 
PSN-IRT remained computationally manageable and showed competitive performance in many settings (except with small sample sizes or in ranking recovery), suggesting that neural estimators may be useful for very large benchmarks.

Sample size requirements depend on the inference target. 
Item analysis claims are less reliable for small model pools. 
\textit{Regardless of capability distribution, $N=30$ was insufficient for recovering item difficulty, discrimination, or mean scores for any estimator examined}, and recovery improved clearly at $N\geq100$. 
This finding cautions against use of item parameters to characterize item difficulty, informativeness or benchmark quality when sample size is very small. 
Model-level aggregate score was less sensitive to sample size when benchmarks were long, as averaging across many items stabilized expected scores. 
Thus, small model pools may support aggregate score prediction but could be insufficient for item-quality analysis or benchmark auditing.

\textit{Model ranking recovery was strongly related to difficulty coverage and capability-distribution skewness}, with poorer recovery when few items had difficulties near the evaluated models’ capabilities or when capabilities were highly skewed. 
\textit{Larger sample sizes and alternative estimators did not resolve these ranking challenges.} 
For ranking claims, researchers may consider running simulation checks using their estimated capability and item parameters as a true data-generating model to assess whether rankings are recoverable under the observed sample size, item pool and capability distribution.

\section{Limitations and Conclusion}\label{sec:7}
This paper investigated the reliability of IRT when applied under the data regimes typical of AI benchmarking. 
Through a structured review of existing studies, we identified substantial heterogeneity in IRT model choices, estimation methods, and data conditions, with many studies operating in regimes far from those for which IRT was originally validated. 
Our simulation study, spanning 360 conditions and 18,000 runs across six benchmarks, three IRT models, five sample sizes, and four estimators, revealed several findings with practical implications.

Several limitations remain. 
Computational feasibility was not assessed on uniform hardware, and convergence monitoring was limited to MML-EM. We also did not examine sample sizes between $30$ and $100$, where item parameter recovery improved substantially, or models beyond unidimensional 1, 2, and 3PL, or estimators beyond the four evaluated. 
The conditions under which IRT-based selection improves short-form ranking recovery also require further study. 
Capability skewness and insufficient difficulty coverage were also moderately correlated in the conditions calibrated from the OpenLLM data, which did not allow fully isolating their effects on ranking recovery. 

Our simulations were calibrated to six OpenLLM Leaderboard benchmarks, with conditions informed by the reviewed studies. 
This empirical grounding supports their relevance to similar applications, but estimator comparisons and sample-size guidance remain conditional on the settings examined. 
Therefore, \textbf{when assessing the reliability of IRT-based inferences in a particular application, we suggest researchers to evaluate recovery through simulations based on their own data}: 
Specifically, they can use estimated item and capability parameters to simulate responses under the same sample size and IRT model, refit the model, and evaluate recovery of the quantities of interest. 
This provides a best-case assessment of reliability of inferences assuming the fitted IRT model is correctly specified. 
We release our simulation and evaluation source code to facilitate this practice.

\begin{ack}
    This work was supported by Schmidt Sciences. 
\end{ack}
\bibliographystyle{abbrv}
\bibliography{ref}

@article{morris2019using,
  title={Using simulation studies to evaluate statistical methods},
  author={Morris, Tim P and White, Ian R and Crowther, Michael J},
  journal={Statistics in Medicine},
  volume={38},
  number={11},
  pages={2074--2102},
  year={2019},
  doi={10.1002/sim.8086},
  publisher={John Wiley \& Sons}
}

@article{siepe2024simulation,
  author    = {Siepe, Bj{\"o}rn S. and Barto{\v{s}}, Franti{\v{s}}ek and Morris, Tim P. and Boulesteix, Anne-Laure and Heck, Daniel W. and Pawel, Samuel},
  title     = {Simulation studies for methodological research in psychology: A standardized template for planning, preregistration, and reporting},
  journal   = {Psychological Methods},
  year      = {2024},
  doi       = {10.1037/met0000695},
  publisher = {American Psychological Association}
}

@ARTICLE{10.3389/fpsyg.2016.00109,
    
AUTHOR={Jiang, Shengyu  and Wang, Chun  and Weiss, David J },
           
TITLE={Sample Size Requirements for Estimation of Item Parameters in the Multidimensional Graded Response Model},
          
JOURNAL={Frontiers in Psychology},
          
VOLUME={Volume 7 - 2016},
  
YEAR={2016},
  
URL={https://www.frontiersin.org/journals/psychology/articles/10.3389/fpsyg.2016.00109},
  
DOI={10.3389/fpsyg.2016.00109},
  
ISSN={1664-1078}}

@book{de2013theory,
  title={The theory and practice of item response theory},
  author={De Ayala, Rafael Jaime},
  year={2013},
  publisher={Guilford Publications}
}

@ARTICLE{Sen2023-ye,
  title     = "The impact of sample size and various other factors on estimation
               of dichotomous mixture {IRT} models",
  author    = "Sen, Sedat and Cohen, Allan S",
  journal   = "Educ. Psychol. Meas.",
  publisher = "SAGE Publications",
  volume    =  83,
  number    =  3,
  pages     = "520--555",
  month     =  jun,
  year      =  2023,
  language  = "en"
}

@article{schroeders2025sample,
  author    = {Schroeders, Ulrich and Gnambs, Timo},
  title     = {Sample-size planning in item-response theory: A tutorial},
  journal   = {Advances in Methods and Practices in Psychological Science},
  year      = {2025},
  volume    = {8},
  number    = {1},
  publisher = {Sage Publications},
  doi       = {10.1177/25152459251314798}
}

@book{demars2010item,
  author    = {DeMars, Christine},
  title     = {Item Response Theory},
  year      = {2010},
  publisher = {Oxford University Press},
  address   = {New York, NY},
  series    = {Understanding Statistics: Measurement},
  isbn      = {978-0-19-537703-3},
  doi       = {10.1093/acprof:oso/9780195377033.001.0001}
}

@article{stone1992recovery,
  author    = {Stone, Clement A.},
  title     = {Recovery of marginal maximum likelihood estimates in the two-parameter logistic response model: An evaluation of {MULTILOG}},
  journal   = {Applied Psychological Measurement},
  year      = {1992},
  volume    = {16},
  number    = {1},
  pages     = {1--16},
  publisher = {Sage Publications},
  doi       = {10.1177/014662169201600101}
}

@article{woods2006item,
  author    = {Woods, Carol M. and Thissen, David},
  title     = {Item response theory with estimation of the latent population distribution using spline-based densities},
  journal   = {Psychometrika},
  year      = {2006},
  volume    = {71},
  number    = {2},
  pages     = {281--301},
  publisher = {Springer},
  doi       = {10.1007/s11336-004-1175-8}
}

@article{kirisci2001robustness,
  author    = {Kirisci, Levent and Hsu, Tse-chi and Yu, Lifa},
  title     = {Robustness of item parameter estimation programs to assumptions of unidimensionality and normality},
  journal   = {Applied Psychological Measurement},
  year      = {2001},
  volume    = {25},
  number    = {2},
  pages     = {146--162},
  publisher = {Sage Publications},
  doi       = {10.1177/01466210122031975}
}

@inproceedings{
hofmann2025fluid,
title={Fluid Language Model Benchmarking},
author={Valentin Hofmann and David Heineman and Ian Magnusson and Kyle Lo and Jesse Dodge and Maarten Sap and Pang Wei Koh and Chun Wang and Hannaneh Hajishirzi and Noah A. Smith},
booktitle={Second Conference on Language Modeling},
year={2025},
url={https://openreview.net/forum?id=mxcCg9YRqj}
}

@inproceedings{ethayarajh-jurafsky-2020-utility,
    title = "Utility is in the Eye of the User: A Critique of {NLP} Leaderboards",
    author = "Ethayarajh, Kawin  and
      Jurafsky, Dan",
    editor = "Webber, Bonnie  and
      Cohn, Trevor  and
      He, Yulan  and
      Liu, Yang",
    booktitle = "Proceedings of the 2020 Conference on Empirical Methods in Natural Language Processing (EMNLP)",
    month = nov,
    year = "2020",
    address = "Online",
    publisher = "Association for Computational Linguistics",
    url = "https://aclanthology.org/2020.emnlp-main.393/",
    doi = "10.18653/v1/2020.emnlp-main.393",
    pages = "4846--4853"
}

@inproceedings{gill-etal-2025-lost,
    title = "What Has Been Lost with Synthetic Evaluation?",
    author = "Gill, Alexander  and
      Ravichander, Abhilasha  and
      Marasovic, Ana",
    editor = "Christodoulopoulos, Christos  and
      Chakraborty, Tanmoy  and
      Rose, Carolyn  and
      Peng, Violet",
    booktitle = "Findings of the Association for Computational Linguistics: EMNLP 2025",
    month = nov,
    year = "2025",
    address = "Suzhou, China",
    publisher = "Association for Computational Linguistics",
    url = "https://aclanthology.org/2025.findings-emnlp.526/",
    doi = "10.18653/v1/2025.findings-emnlp.526",
    pages = "9902--9945",
    ISBN = "979-8-89176-335-7"
}

@inproceedings{rodriguez-etal-2021-evaluation,
    title = "Evaluation Examples are not Equally Informative: How should that change {NLP} Leaderboards?",
    author = "Rodriguez, Pedro  and
      Barrow, Joe  and
      Hoyle, Alexander  and
      Lalor, John P.  and
      Jia, Robin  and
      Boyd-Graber, Jordan",
    editor = "Zong, Chengqing  and
      Xia, Fei  and
      Li, Wenjie  and
      Navigli, Roberto",
    booktitle = "Proceedings of the 59th Annual Meeting of the Association for Computational Linguistics and the 11th International Joint Conference on Natural Language Processing (Volume 1: Long Papers)",
    month = aug,
    year = "2021",
    address = "Online",
    publisher = "Association for Computational Linguistics",
    url = "https://aclanthology.org/2021.acl-long.346/",
    doi = "10.18653/v1/2021.acl-long.346",
    pages = "4486--4503"
}

@inproceedings{jung2026defines,
  title={Who Defines “Best”? Towards Interactive, User-Defined Evaluation of LLM Leaderboards},
  author={Jung, Minji and Lee, Minjae and Kim, Yejin and Choi, Sarang and Kahng, Minsuk},
  booktitle={The 2026 ACM Conference on Fairness, Accountability, and Transparency},
  pages={4733--4755},
  year={2026}
}

@article{salaudeen2025measurement,
  title={Measurement to meaning: A validity-centered framework for ai evaluation},
  author={Salaudeen, Olawale and Reuel, Anka and Ahmed, Ahmed and Bedi, Suhana and Robertson, Zachary and Sundar, Sudharsan and Domingue, Ben and Wang, Angelina and Koyejo, Sanmi},
  journal={arXiv preprint arXiv:2505.10573},
  year={2025}
}

@inproceedings{
heineman2026signal,
title={Signal and Noise: A Framework for Reducing Uncertainty in Language Model Evaluation},
author={David Heineman and Valentin Hofmann and Ian Magnusson and Yuling Gu and Noah A. Smith and Hannaneh Hajishirzi and Kyle Lo and Jesse Dodge},
booktitle={The Thirty-ninth Annual Conference on Neural Information Processing Systems},
year={2026},
url={https://openreview.net/forum?id=sAFottNlra}
}

@article{ott2022benchmark,
  title   = {Mapping global dynamics of benchmark creation and saturation in artificial intelligence},
  author  = {Ott, Simon and Barbosa-Silva, Adriano and Blagec, Kathrin and Brauner, Jan and Samwald, Matthias},
  journal = {Nature Communications},
  volume  = {13},
  number  = {1},
  pages   = {6793},
  year    = {2022},
  doi     = {10.1038/s41467-022-34591-0},
  issn    = {2041-1723}
}

@inproceedings{
deveci2025the,
title={The Ouroboros of Benchmarking: Reasoning Evaluation in an Era of Saturation},
author={{\.I}brahim Ethem Deveci and Duygu Ataman},
booktitle={NeurIPS 2025 Workshop on Evaluating the Evolving LLM Lifecycle: Benchmarks, Emergent Abilities, and Scaling},
year={2025},
url={https://openreview.net/forum?id=0zDiyIGCFT}
}

@article{
liang2023holistic,
title={Holistic Evaluation of Language Models},
author={Percy Liang and Rishi Bommasani and Tony Lee and Dimitris Tsipras and Dilara Soylu and Michihiro Yasunaga and Yian Zhang and Deepak Narayanan and Yuhuai Wu and Ananya Kumar and Benjamin Newman and Binhang Yuan and Bobby Yan and Ce Zhang and Christian Cosgrove and Christopher D Manning and Christopher Re and Diana Acosta-Navas and Drew A. Hudson and Eric Zelikman and Esin Durmus and Faisal Ladhak and Frieda Rong and Hongyu Ren and Huaxiu Yao and Jue WANG and Keshav Santhanam and Laurel Orr and Lucia Zheng and Mert Yuksekgonul and Mirac Suzgun and Nathan Kim and Neel Guha and Niladri S. Chatterji and Omar Khattab and Peter Henderson and Qian Huang and Ryan Andrew Chi and Sang Michael Xie and Shibani Santurkar and Surya Ganguli and Tatsunori Hashimoto and Thomas Icard and Tianyi Zhang and Vishrav Chaudhary and William Wang and Xuechen Li and Yifan Mai and Yuhui Zhang and Yuta Koreeda},
journal={Transactions on Machine Learning Research},
issn={2835-8856},
year={2023},
url={https://openreview.net/forum?id=iO4LZibEqW},
note={Featured Certification, Expert Certification, Outstanding Certification}
}

@inproceedings{lalor-etal-2016-building,
    title = "Building an Evaluation Scale using Item Response Theory",
    author = "Lalor, John P.  and
      Wu, Hao  and
      Yu, Hong",
    editor = "Su, Jian  and
      Duh, Kevin  and
      Carreras, Xavier",
    booktitle = "Proceedings of the 2016 Conference on Empirical Methods in Natural Language Processing",
    month = nov,
    year = "2016",
    address = "Austin, Texas",
    publisher = "Association for Computational Linguistics",
    url = "https://aclanthology.org/D16-1062/",
    doi = "10.18653/v1/D16-1062",
    pages = "648--657"
}

@inproceedings{vania-etal-2021-comparing,
    title = "Comparing Test Sets with Item Response Theory",
    author = "Vania, Clara  and
      Htut, Phu Mon  and
      Huang, William  and
      Mungra, Dhara  and
      Pang, Richard Yuanzhe  and
      Phang, Jason  and
      Liu, Haokun  and
      Cho, Kyunghyun  and
      Bowman, Samuel R.",
    editor = "Zong, Chengqing  and
      Xia, Fei  and
      Li, Wenjie  and
      Navigli, Roberto",
    booktitle = "Proceedings of the 59th Annual Meeting of the Association for Computational Linguistics and the 11th International Joint Conference on Natural Language Processing (Volume 1: Long Papers)",
    month = aug,
    year = "2021",
    address = "Online",
    publisher = "Association for Computational Linguistics",
    url = "https://aclanthology.org/2021.acl-long.92/",
    doi = "10.18653/v1/2021.acl-long.92",
    pages = "1141--1158"
}

@inproceedings{
schilling-wilhelmi2025lifting,
title={Lifting the benchmark iceberg with item-response theory},
author={Mara Schilling-Wilhelmi and Nawaf Alampara and Kevin Maik Jablonka},
booktitle={AI for Accelerated Materials Design - ICLR 2025},
year={2025},
url={https://openreview.net/forum?id=ZyVQqK7mcP}
}

@book{lord1980applications,
  title={Applications of Item Response Theory to Practical Testing Problems},
  author={Lord, Frederic M.},
  year={1980},
  publisher={Lawrence Erlbaum Associates},
  address={Hillsdale, NJ},
  isbn={089859006X}
}

@book{hambleton1991fundamentals,
  title={Fundamentals of Item Response Theory},
  author={Hambleton, Ronald K. and Swaminathan, Hariharan and Rogers, H. Jane},
  year={1991},
  publisher={Sage Publications},
  address={Thousand Oaks, CA},
  series={Measurement Methods for the Social Sciences},
  volume={2},
  isbn={0-8039-3647-8}
}

@book{lord1968statistical,
  title={Statistical Theories of Mental Test Scores},
  author={Lord, Frederic M. and Novick, Melvin R.},
  year={1968},
  publisher={Addison-Wesley},
  address={Reading, MA}
}

@book{rasch1960probabilistic,
  author    = {Rasch, Georg},
  title     = {Probabilistic Models for Some Intelligence and Attainment Tests},
  year      = {1960},
  publisher = {Nielsen \& Lydiche},
  address   = {Copenhagen, Denmark}
}

@inproceedings{Lord1966SomeLT,
  title={Some latent trait models and their use in inferring an examinee's ability},
  author={Frederic M. Lord and Melvin R. Novick and Allan Birnbaum},
  year={1966},
  url={https://api.semanticscholar.org/CorpusID:61082238}
}

@article{4pl,
author = {Barton, Mark A. and Lord, Frederic M.},
title = {AN UPPER ASYMPTOTE FOR THE THREE-PARAMETER LOGISTIC ITEM-RESPONSE MODEL},
journal = {ETS Research Report Series},
volume = {1981},
number = {1},
pages = {i-8},
doi = {https://doi.org/10.1002/j.2333-8504.1981.tb01255.x},
url = {https://onlinelibrary.wiley.com/doi/abs/10.1002/j.2333-8504.1981.tb01255.x},
eprint = {https://onlinelibrary.wiley.com/doi/pdf/10.1002/j.2333-8504.1981.tb01255.x},
year = {1981}
}

@article{samejima1969estimation,
  author  = {Samejima, Fumiko},
  title   = {Estimation of Latent Ability Using a Response Pattern of Graded Scores},
  journal = {Psychometrika},
  year    = {1969},
  volume  = {34},
  number  = {4, Pt. 2},
  pages   = {1--97},
  doi     = {10.1007/BF03372160},
  note    = {Psychometrika Monograph Supplement No.~17}
}

@article{bock1981marginal,
  author  = {Bock, R. Darrell and Aitkin, Murray},
  title   = {Marginal Maximum Likelihood Estimation of Item Parameters: Application of an {EM} Algorithm},
  journal = {Psychometrika},
  year    = {1981},
  volume  = {46},
  number  = {4},
  pages   = {443--459},
  doi     = {10.1007/BF02293801}
}

@article{mcmc1,
 ISSN = {03629791},
 URL = {http://www.jstor.org/stable/1165149},
 author = {James H. Albert},
 journal = {Journal of Educational Statistics},
 number = {3},
 pages = {251--269},
 publisher = {[Sage Publications, Inc., American Educational Research Association, American Statistical Association]},
 title = {Bayesian Estimation of Normal Ogive Item Response Curves Using Gibbs Sampling},
 urldate = {2026-05-02},
 volume = {17},
 year = {1992}
}

@article{mcmc2,
 ISSN = {01621459, 1537274X},
 URL = {http://www.jstor.org/stable/2290350},
 author = {James H. Albert and Siddhartha Chib},
 journal = {Journal of the American Statistical Association},
 number = {422},
 pages = {669--679},
 publisher = {[American Statistical Association, Taylor & Francis, Ltd.]},
 title = {Bayesian Analysis of Binary and Polychotomous Response Data},
 urldate = {2026-05-02},
 volume = {88},
 year = {1993}
}

@misc{wu2020variationalitemresponsetheory,
      title={Variational Item Response Theory: Fast, Accurate, and Expressive}, 
      author={Mike Wu and Richard L. Davis and Benjamin W. Domingue and Chris Piech and Noah Goodman},
      year={2020},
      eprint={2002.00276},
      archivePrefix={arXiv},
      primaryClass={cs.LG},
      url={https://arxiv.org/abs/2002.00276}, 
}

@InProceedings{pmlr-v238-frick24a,
  title = 	 {Scalable Learning of Item Response Theory Models},
  author =       {Frick, Susanne and Krivosija, Amer and Munteanu, Alexander},
  booktitle = 	 {Proceedings of The 27th International Conference on Artificial Intelligence and Statistics},
  pages = 	 {1234--1242},
  year = 	 {2024},
  editor = 	 {Dasgupta, Sanjoy and Mandt, Stephan and Li, Yingzhen},
  volume = 	 {238},
  series = 	 {Proceedings of Machine Learning Research},
  month = 	 {02--04 May},
  publisher =    {PMLR},
  url = 	 {https://proceedings.mlr.press/v238/frick24a.html}
}

@article{lost, title={Lost in Benchmarks? Rethinking Large Language Model Benchmarking with Item Response Theory}, volume={40}, url={https://ojs.aaai.org/index.php/AAAI/article/view/40814}, DOI={10.1609/aaai.v40i41.40814}, abstractNote={The evaluation of large language models (LLMs) via benchmarks is widespread, yet inconsistencies between different leaderboards and poor separability among top models raise concerns about their ability to accurately reflect authentic model capabilities. This paper provides a critical analysis of benchmark effectiveness, examining mainstream prominent LLM benchmarks using results from diverse models. We first propose Pseudo-Siamese Network for Item Response Theory (PSN-IRT), an enhanced Item Response Theory framework that incorporates a rich set of item parameters within an IRT-grounded architecture. PSN-IRT can be utilized for accurate and reliable estimations of item characteristics and model abilities. Based on PSN-IRT, we conduct extensive analysis on 11 LLM benchmarks comprising 41,871 items, revealing significant and varied shortcomings in their measurement quality. Furthermore, we demonstrate that leveraging PSN-IRT is able to construct smaller benchmarks while maintaining stronger alignment with human preference.}, number={41}, journal={Proceedings of the AAAI Conference on Artificial Intelligence}, author={Zhou, Hongli and Huang, Hui and Zhao, Ziqing and Han, Lvyuan and Wang, Huicheng and Chen, Kehai and Yang, Muyun and Bao, Wei and Dong, Jian and Xu, Bing and Zhu, Conghui and Cao, Hailong and Zhao, Tiejun}, year={2026}, month={Mar.}, pages={35085-35093} }

@inproceedings{siska-etal-2024-examining,
    title = "Examining the robustness of {LLM} evaluation to the distributional assumptions of benchmarks",
    author = "Siska, Charlotte  and
      Marazopoulou, Katerina  and
      Ailem, Melissa  and
      Bono, James",
    editor = "Ku, Lun-Wei  and
      Martins, Andre  and
      Srikumar, Vivek",
    booktitle = "Proceedings of the 62nd Annual Meeting of the Association for Computational Linguistics (Volume 1: Long Papers)",
    month = aug,
    year = "2024",
    address = "Bangkok, Thailand",
    publisher = "Association for Computational Linguistics",
    url = "https://aclanthology.org/2024.acl-long.560/",
    doi = "10.18653/v1/2024.acl-long.560",
    pages = "10406--10421"
}

@ARTICLE{Xu2025-ye,
  title         = "Latency-Response Theory model: Evaluating Large Language
                   Models via response accuracy and chain-of-thought length",
  author        = "Xu, Zhiyu and Liu, Jia and Wang, Yixin and Gu, Yuqi",
  journal       = "arXiv [stat.ME]",
  month         =  dec,
  year          =  2025,
  archivePrefix = "arXiv",
  primaryClass  = "stat.ME"
}

@ARTICLE{Robertson2025-ld,
  title         = "Identity-link {IRT} for label-free {LLM} evaluation:
                   Preserving additivity in {TVD}-{MI} scores",
  author        = "Robertson, Zachary",
  journal       = "arXiv [cs.LG]",
  month         =  oct,
  year          =  2025,
  archivePrefix = "arXiv",
  primaryClass  = "cs.LG"
}

@ARTICLE{Liao2025-wn,
  title         = "Toward a unified framework for data-efficient evaluation of
                   large language models",
  author        = "Liao, Lele and Zhang, Qile and Wu, Ruofan and Fang, Guanhua",
  journal       = "arXiv [cs.AI]",
  month         =  oct,
  year          =  2025,
  archivePrefix = "arXiv",
  primaryClass  = "cs.AI"
}

@ARTICLE{Chen2025-aa,
  title         = "Learning compact representations of {LLM} abilities via item
                   response theory",
  author        = "Chen, Jianhao and Wang, Chenxu and Zhang, Gengrui and Ye,
                   Peng and Bai, Lei and Hu, Wei and Qu, Yuzhong and Hu, Shuyue",
  journal       = "arXiv [cs.AI]",
  month         =  oct,
  year          =  2025,
  archivePrefix = "arXiv",
  primaryClass  = "cs.AI"
}

@INPROCEEDINGS{Jiang2025-jc,
  title     = "Raising the Bar: Investigating the Values of Large Language
               Models via Generative Evolving Testing",
  author    = "Jiang, Han and Yi, Xiaoyuan and Wei, Zhihua and Xiao, Ziang and
               Wang, Shu and Xie, Xing",
  booktitle = "Forty-second International Conference on Machine Learning",
  month     =  jun,
  year      =  2025
}

@INPROCEEDINGS{Truong2025-zj,
  title     = "Reliable and Efficient Amortized Model-based Evaluation",
  author    = "Truong, Sang T and Tu, Yuheng and Liang, Percy and Li, Bo and
               Koyejo, Sanmi",
  booktitle = "Forty-second International Conference on Machine Learning",
  month     =  jun,
  year      =  2025
}

@INPROCEEDINGS{Kipnis2024-ex,
  title     = "metabench - A Sparse Benchmark of Reasoning and Knowledge in
               Large Language Models",
  author    = "Kipnis, Alex and Voudouris, Konstantinos and Schulze Buschoff,
               Luca M and Schulz, Eric",
  booktitle = "The Thirteenth International Conference on Learning
               Representations",
  month     =  oct,
  year      =  2024
}

@INPROCEEDINGS{Polo2024-mu,
  title     = "{tinyBenchmarks}: evaluating {LLMs} with fewer examples",
  author    = "Polo, Felipe Maia and Weber, Lucas and Choshen, Leshem and Sun,
               Yuekai and Xu, Gongjun and Yurochkin, Mikhail",
  booktitle = "International Conference on Machine Learning",
  publisher = "PMLR",
  pages     = "34303--34326",
  month     =  jul,
  year      =  2024,
  language  = "en"
}

@MISC{Zhuang2023-kk,
  title        = "Efficiently Measuring the Cognitive Ability of {LLMs}: An
                  Adaptive Testing Perspective",
  author       = "Zhuang, Yan and Liu, Qi and Ning, Yuting and Huang, Weizhe and
                  Lv, Rui and Huang, Zhenya and Zhao, Guanhao and Zhang, Zheng
                  and Mao, Qingyang and Wang, Shijin and Chen, Enhong",
  month        =  oct,
  year         =  2023,
  howpublished = "\url{https://openreview.net/forum?id=s6X3s3rBPW}",
  note         = "Accessed: 2026-4-16",
  language     = "en"
}

@INPROCEEDINGS{Byrd2022-cq,
  title     = "Predicting difficulty and discrimination of natural language
               questions",
  author    = "Byrd, Matthew and Srivastava, Shashank",
  booktitle = "Proceedings of the 60th Annual Meeting of the Association for
               Computational Linguistics (Volume 2: Short Papers)",
  publisher = "Association for Computational Linguistics",
  address   = "Stroudsburg, PA, USA",
  pages     = "119--130",
  year      =  2022
}

@ARTICLE{Lalor2019-uw,
  title     = "Learning latent parameters without human response patterns: Item
               Response Theory with artificial crowds",
  author    = "Lalor, John P and Wu, Hao and Yu, Hong",
  journal   = "Proc. Conf. Empir. Methods Nat. Lang. Process.",
  publisher = "Association for Computational Linguistics",
  address   = "Stroudsburg, PA, USA",
  volume    =  2019,
  pages     = "4240--4250",
  month     =  nov,
  year      =  2019,
  language  = "en"
}

@ARTICLE{Martinez-Plumed2019-jv,
  title     = "Item response theory in {AI}: Analysing machine learning
               classifiers at the instance level",
  author    = "Martínez-Plumed, Fernando and Prudêncio, Ricardo B C and
               Martínez-Usó, Adolfo and Hernández-Orallo, José",
  journal   = "Artif. Intell.",
  publisher = "Elsevier BV",
  volume    =  271,
  pages     = "18--42",
  month     =  jun,
  year      =  2019,
  language  = "en"
}

@ARTICLE{Lalor2018-gn,
  title     = "Understanding deep learning performance through an examination of
               test set difficulty: A psychometric case study",
  author    = "Lalor, John P and Wu, Hao and Munkhdalai, Tsendsuren and Yu, Hong",
  journal   = "Proc. Conf. Empir. Methods Nat. Lang. Process.",
  publisher = "Association for Computational Linguistics",
  address   = "Stroudsburg, PA, USA",
  volume    =  2018,
  pages     = "4711--4716",
  month     =  oct,
  year      =  2018,
  language  = "en"
}

@misc{jiang2026position,
      title={Position: Science of AI Evaluation Requires Item-level Benchmark Data}, 
      author={Han Jiang and Susu Zhang and Xiaoyuan Yi and Xing Xie and Ziang Xiao},
      year={2026},
      eprint={2604.03244},
      archivePrefix={arXiv},
      primaryClass={cs.AI},
      url={https://arxiv.org/abs/2604.03244}, 
}

@article{pyirt,
author = {Lalor, John Patrick and Rodriguez, Pedro},
title = {py-irt: A Scalable Item Response Theory Library for Python},
year = {2023},
issue_date = {January-February 2023},
publisher = {INFORMS},
address = {Linthicum, MD, USA},
volume = {35},
number = {1},
issn = {1526-5528},
url = {https://doi.org/10.1287/ijoc.2022.1250},
doi = {10.1287/ijoc.2022.1250},
journal = {INFORMS J. on Computing},
month = jan,
pages = {5–13},
numpages = {9}
}

@article{hambleton1993comparison,
  title={Comparison of classical test theory and item response theory and their applications to test development},
  author={Hambleton, Ronald K. and Jones, Russell W.},
  journal={Educational Measurement: Issues and Practice},
  volume={12},
  number={3},
  pages={38--47},
  year={1993},
  publisher={Blackwell Publishing},
  doi={10.1111/j.1745-3992.1993.tb00543.x}
}

@book{gulliksen1950theory,
  author    = {Gulliksen, Harold},
  title     = {Theory of Mental Tests},
  publisher = {John Wiley \& Sons},
  address   = {Hoboken, NJ},
  year      = {1950},
  series    = {Wiley Publications in Psychology},
  pages     = {xix--486},
  doi       = {10.1037/13240-000}
}

@misc{myrzakhan2024openllm,
      title={Open-LLM-Leaderboard: From Multi-choice to Open-style Questions for LLMs Evaluation, Benchmark, and Arena}, 
      author={Aidar Myrzakhan and Sondos Mahmoud Bsharat and Zhiqiang Shen},
      year={2024},
      eprint={2406.07545},
      archivePrefix={arXiv},
      primaryClass={cs.CL},
      url={https://arxiv.org/abs/2406.07545}, 
}

@misc{clark2018arc,
      title={Think you have Solved Question Answering? Try ARC, the AI2 Reasoning Challenge}, 
      author={Peter Clark and Isaac Cowhey and Oren Etzioni and Tushar Khot and Ashish Sabharwal and Carissa Schoenick and Oyvind Tafjord},
      year={2018},
      eprint={1803.05457},
      archivePrefix={arXiv},
      primaryClass={cs.AI},
      url={https://arxiv.org/abs/1803.05457}, 
}

@inproceedings{zellers-etal-2019-hellaswag,
    title = "{H}ella{S}wag: Can a Machine Really Finish Your Sentence?",
    author = "Zellers, Rowan  and
      Holtzman, Ari  and
      Bisk, Yonatan  and
      Farhadi, Ali  and
      Choi, Yejin",
    editor = "Korhonen, Anna  and
      Traum, David  and
      M{\`a}rquez, Llu{\'i}s",
    booktitle = "Proceedings of the 57th Annual Meeting of the Association for Computational Linguistics",
    month = jul,
    year = "2019",
    address = "Florence, Italy",
    publisher = "Association for Computational Linguistics",
    url = "https://aclanthology.org/P19-1472/",
    doi = "10.18653/v1/P19-1472",
    pages = "4791--4800"
}

@article{hendryckstest2021,
  title={Measuring Massive Multitask Language Understanding},
  author={Dan Hendrycks and Collin Burns and Steven Basart and Andy Zou and Mantas Mazeika and Dawn Song and Jacob Steinhardt},
  journal={Proceedings of the International Conference on Learning Representations (ICLR)},
  year={2021}
}

@inproceedings{lin-etal-2022-truthfulqa,
    title = "{T}ruthful{QA}: Measuring How Models Mimic Human Falsehoods",
    author = "Lin, Stephanie  and
      Hilton, Jacob  and
      Evans, Owain",
    editor = "Muresan, Smaranda  and
      Nakov, Preslav  and
      Villavicencio, Aline",
    booktitle = "Proceedings of the 60th Annual Meeting of the Association for Computational Linguistics (Volume 1: Long Papers)",
    month = may,
    year = "2022",
    address = "Dublin, Ireland",
    publisher = "Association for Computational Linguistics",
    url = "https://aclanthology.org/2022.acl-long.229/",
    doi = "10.18653/v1/2022.acl-long.229",
    pages = "3214--3252"
}

@article{10.1145/3474381,
author = {Sakaguchi, Keisuke and Bras, Ronan Le and Bhagavatula, Chandra and Choi, Yejin},
title = {WinoGrande: an adversarial winograd schema challenge at scale},
year = {2021},
issue_date = {September 2021},
publisher = {Association for Computing Machinery},
address = {New York, NY, USA},
volume = {64},
number = {9},
issn = {0001-0782},
url = {https://doi.org/10.1145/3474381},
doi = {10.1145/3474381},
journal = {Commun. ACM},
month = aug,
pages = {99–106},
numpages = {8}
}

@misc{cobbe2021gsm8k,
      title={Training Verifiers to Solve Math Word Problems}, 
      author={Karl Cobbe and Vineet Kosaraju and Mohammad Bavarian and Mark Chen and Heewoo Jun and Lukasz Kaiser and Matthias Plappert and Jerry Tworek and Jacob Hilton and Reiichiro Nakano and Christopher Hesse and John Schulman},
      year={2021},
      eprint={2110.14168},
      archivePrefix={arXiv},
      primaryClass={cs.LG},
      url={https://arxiv.org/abs/2110.14168}, 
}

@article{burkner2021bayesian,
  title={Bayesian item response modeling in R with brms and Stan},
  author={B{\"u}rkner, Paul-Christian},
  journal={Journal of statistical software},
  volume={100},
  pages={1--54},
  year={2021}
}

@article{skewness,
 ISSN = {00390526, 14679884},
 URL = {http://www.jstor.org/stable/2988433},
 author = {D. N. Joanes and C. A. Gill},
 journal = {Journal of the Royal Statistical Society. Series D (The Statistician)},
 number = {1},
 pages = {183--189},
 publisher = {[Royal Statistical Society, Wiley]},
 title = {Comparing Measures of Sample Skewness and Kurtosis},
 urldate = {2026-05-06},
 volume = {47},
 year = {1998}
}

@article{mirt,
 title={mirt: A Multidimensional Item Response Theory Package for the R Environment},
 volume={48},
 url={https://www.jstatsoft.org/index.php/jss/article/view/v048i06},
 doi={10.18637/jss.v048.i06},
 number={6},
 journal={Journal of Statistical Software},
 author={Chalmers, R. Philip},
 year={2012},
 pages={1–29}
}

@article{brms,
 title={brms: An R Package for Bayesian Multilevel Models Using Stan},
 volume={80},
 url={https://www.jstatsoft.org/index.php/jss/article/view/v080i01},
 doi={10.18637/jss.v080.i01},
 number={1},
 journal={Journal of Statistical Software},
 author={Bürkner, Paul-Christian},
 year={2017},
 pages={1–28}
}

@misc{bommasani2022opportunities,
      title={On the Opportunities and Risks of Foundation Models}, 
      author={Rishi Bommasani and Drew A. Hudson and Ehsan Adeli and Russ Altman and Simran Arora and Sydney von Arx and Michael S. Bernstein and Jeannette Bohg and Antoine Bosselut and Emma Brunskill and Erik Brynjolfsson and Shyamal Buch and Dallas Card and Rodrigo Castellon and Niladri Chatterji and Annie Chen and Kathleen Creel and Jared Quincy Davis and Dora Demszky and Chris Donahue and Moussa Doumbouya and Esin Durmus and Stefano Ermon and John Etchemendy and Kawin Ethayarajh and Li Fei-Fei and Chelsea Finn and Trevor Gale and Lauren Gillespie and Karan Goel and Noah Goodman and Shelby Grossman and Neel Guha and Tatsunori Hashimoto and Peter Henderson and John Hewitt and Daniel E. Ho and Jenny Hong and Kyle Hsu and Jing Huang and Thomas Icard and Saahil Jain and Dan Jurafsky and Pratyusha Kalluri and Siddharth Karamcheti and Geoff Keeling and Fereshte Khani and Omar Khattab and Pang Wei Koh and Mark Krass and Ranjay Krishna and Rohith Kuditipudi and Ananya Kumar and Faisal Ladhak and Mina Lee and Tony Lee and Jure Leskovec and Isabelle Levent and Xiang Lisa Li and Xuechen Li and Tengyu Ma and Ali Malik and Christopher D. Manning and Suvir Mirchandani and Eric Mitchell and Zanele Munyikwa and Suraj Nair and Avanika Narayan and Deepak Narayanan and Ben Newman and Allen Nie and Juan Carlos Niebles and Hamed Nilforoshan and Julian Nyarko and Giray Ogut and Laurel Orr and Isabel Papadimitriou and Joon Sung Park and Chris Piech and Eva Portelance and Christopher Potts and Aditi Raghunathan and Rob Reich and Hongyu Ren and Frieda Rong and Yusuf Roohani and Camilo Ruiz and Jack Ryan and Christopher Ré and Dorsa Sadigh and Shiori Sagawa and Keshav Santhanam and Andy Shih and Krishnan Srinivasan and Alex Tamkin and Rohan Taori and Armin W. Thomas and Florian Tramèr and Rose E. Wang and William Wang and Bohan Wu and Jiajun Wu and Yuhuai Wu and Sang Michael Xie and Michihiro Yasunaga and Jiaxuan You and Matei Zaharia and Michael Zhang and Tianyi Zhang and Xikun Zhang and Yuhui Zhang and Lucia Zheng and Kaitlyn Zhou and Percy Liang},
      year={2022},
      eprint={2108.07258},
      archivePrefix={arXiv},
      primaryClass={cs.LG},
      url={https://arxiv.org/abs/2108.07258}, 
}

@inproceedings{NEURIPS2023_a0b1082f,
 author = {Toups, Connor and Bommasani, Rishi and Creel, Kathleen and Bana, Sarah and Jurafsky, Dan and Liang, Percy S},
 booktitle = {Advances in Neural Information Processing Systems},
 doi = {10.52202/075280-2228},
 editor = {A. Oh and T. Naumann and A. Globerson and K. Saenko and M. Hardt and S. Levine},
 pages = {51178--51201},
 publisher = {Curran Associates, Inc.},
 title = {Ecosystem-level Analysis of Deployed Machine Learning Reveals Homogeneous Outcomes},
 url = {https://proceedings.neurips.cc/paper_files/paper/2023/file/a0b1082fc7823c4c68abcab4fa850e9c-Paper-Conference.pdf},
 volume = {36},
 year = {2023}
}

@article{bechger2003using,
title={Using Classical Test Theory in Combination with Item Response Theory},
author={Bechger, Timo M. and Maris, Gunter and Verstralen, Huub H. F. M. and B{'e}guin, Anton A.},
journal={Applied Psychological Measurement},
volume={27},
number={5},
pages={319--334},
year={2003},
doi={10.1177/0146621603257518}
}


\appendix
\section{Estimator introduction}\label{app:estimation}

\paragraph{MML-EM.} Marginal maximum likelihood (MML) is the conventional approach to IRT parameter estimation for human testing. It treats latent traits as random draws from a population distribution, often standard normal for computational convenience, and estimates item parameters, ($\boldsymbol{\zeta}_j, j = 1, \ldots, J$) by maximizing the marginal likelihood given the response matrix:
\begin{equation}
L_{MML}(\boldsymbol{\zeta}) =
\prod_{i=1}^{N} \int \left[
\prod_{j=1}^{J} p(Y_{ij} \mid \theta, \boldsymbol{\zeta}_j)
\right]
f(\theta)\, d\theta .
\end{equation}
Optimization is done with the EM algorithm. We used the \texttt{mirt} package and adopted the stochastic variant, the Metropolis-Hastings Robbins-Monro algorithm, replacing fixed quadrature with Monte Carlo integration when standard EM is too costly for large item pools. Statistical consistency theory guarantees the MML estimator to recover true item parameters when the number of items ($J$) is fixed and sample size ($N$) grows to infinity. Although in human testing, often the assumption of trait normality is empirically plausible and $J<N$, MML's behavior is less certain in AI benchmark regimes, where the number of evaluated systems may be small, the number of items large, and the capability distribution nonnormal.

\paragraph{MCMC.} Bayesian MCMC specifies prior distributions for item parameters and model capabilities, and then samples from the posterior distribution after observing the response data ($\mathbf Y$),
\begin{equation}\label{eq:bayes_post}
p(\boldsymbol{\theta},\boldsymbol{\zeta}\mid Y)
\propto
p(Y\mid \boldsymbol{\theta},\boldsymbol{\zeta})
p(\boldsymbol{\theta},\boldsymbol{\zeta}).
\end{equation}
We implemented Bayesian estimation in \texttt{brms}, which uses Stan's Hamiltonian Monte Carlo sampler. HMC uses gradient information to reduce random-walk behavior, although sampling efficiency still depends on posterior geometry and model size. Prior settings and implementation details are reported in Appendix A. Bayesian posterior estimates can be sensitive to prior specification when empirical information is limited, especially when few evaluated systems are available for estimating many item parameters. Further, runtime grows with both sample size and benchmark length as each MCMC iteration evaluates the full likelihood over all model and item parameters. 

The prior specifications for the MCMC algorithm in this study followed \cite{burkner2021bayesian} and are summarized in Table \ref{tab:prior_specifications}. For each model, the MCMC algorithm was run using a single chain with 2,000 total iterations, where the first 1,000 iterations were discarded as burn-in. 

\paragraph{Variational inference.} VI in this study was fitted with \texttt{py-irt}, which approximates the Bayesian posterior in Eq.~\eqref{eq:bayes_post} with a tractable variational distribution, e.g., for the 1PL model,
\begin{equation}\label{eq:vi_dist}
q_{\phi}(\boldsymbol{\theta},\mathbf{b},\mu,\tau)=
q(\mu)q(\tau)\prod_{i=1}^{N}q(\theta_i)\prod_{j=1}^{J}q(b_j),
\end{equation}
where $\mu$ and $\tau$ are the location and precision parameters of the normal latent capability. The variational parameters are estimated by minimizing $D_{\mathrm{KL}}(q_{\phi}\,\|\,p)$, the KL divergence between \eqref{eq:vi_dist} and \eqref{eq:bayes_post}. VI is computationally attractive for long benchmark matrices as it replaces numerical integration or posterior sampling with gradient-based optimization. However, it inherits the prior assumptions of the Bayesian model and adds approximation error from the chosen variational family, which can affect inferences in small-sample AI benchmark settings and when capability distributions is non-normal.

\paragraph{PSN-IRT.} PSN \cite{lost} is a neural-network representation of an IRT model, which assigns each evaluated model a learned capability embedding ($\theta_i$) and each benchmark item item-parameter embeddings $(a_j,b_j,c_j)$. These embeddings are passed through an item response function layer to predict binary response. We adapted the authors' original 4PL implementation to the 1 - 3PL models used in our simulation. PSN is computationally efficient with gradient-based optimization and does not assuming a normal distribution for latent capability. However, it is trained as a predictive neural model and does not directly provide uncertainty quantification or parameter consistency guarantees for parameter estimates, and its estimates are specific to the trained network and item set.

\begin{table}[htbp]
    \centering
    \small
    \caption{Prior Specifications for MCMC Algorithm. $\text{Normal}^+$ denotes a half-normal distribution.}
    \label{tab:prior_specifications}
    \begin{tabular}{llll}
        \toprule
        \textbf{IRT Model} & \textbf{Parameter} & \textbf{Prior} & \textbf{Hyperprior } \\
        \midrule
        \textbf{1PL} 
        & Ability ($\theta$) & $\theta_i\!\sim\!\text{Normal}(0, \sigma_\theta)$ & $\sigma_\theta\!\sim\!\text{Normal}^+(0, 1)$ \\
        \cmidrule{2-4}
        & Difficulty ($b$) & $b_j\!\sim\!\text{Normal}(\mu_b, \sigma_b)$ & $\sigma_b\!\sim\!\text{Normal}^+(0, 3)$ \\
        \midrule
        \textbf{2PL \& 3PL} 
        & Ability ($\theta$) & $\theta_i\!\sim\!\text{Normal}(0, 1)$ & $\sigma_\theta = 1$ (Fixed scale) \\
        \cmidrule{2-4}
        & Difficulty ($b$) & $b_j\!\sim\!\text{Normal}(\mu_b, \sigma_b)$ & \begin{tabular}{@{}l@{}}$\mu_b\!\sim\!\text{Normal}(0, 5)$ \\ $\sigma_b\!\sim\!\text{Normal}^+(0, 3)$\end{tabular} \\
        \cmidrule{2-4}
        & Discrimination ($a$) & $\log a_j\!\sim\!\text{Normal}(\mu_{\log a}, \sigma_{\log a})$ & \begin{tabular}{@{}l@{}}$\mu_{\log a}\!\sim\!\text{Normal}(0, 1)$ \\ $\sigma_{\log a}\!\sim\!\text{Normal}^+(0, 1)$\end{tabular} \\
        \midrule
        \textbf{3PL} 
        & Pseudo-Guessing ($c$) & $\text{logit } c_j\!\sim\!\text{Normal}(\mu_{\text{logit } c}, \sigma_{\text{logit } c})$ & \begin{tabular}{@{}l@{}}$\mu_{\text{logit } c}\!\sim\!\text{Logistic}(0, 1)$ \\ $\sigma_{\text{logit } c}\!\sim\!\text{Normal}^+(0, 1)$\end{tabular} \\
        \bottomrule
    \end{tabular}
\end{table}

\section{Data-generating parameters}
\subsection{True capability and difficulty distributions}\label{app:theta_beta}
\begin{figure*}[ht]
    \centering
    \includegraphics[width=\linewidth]{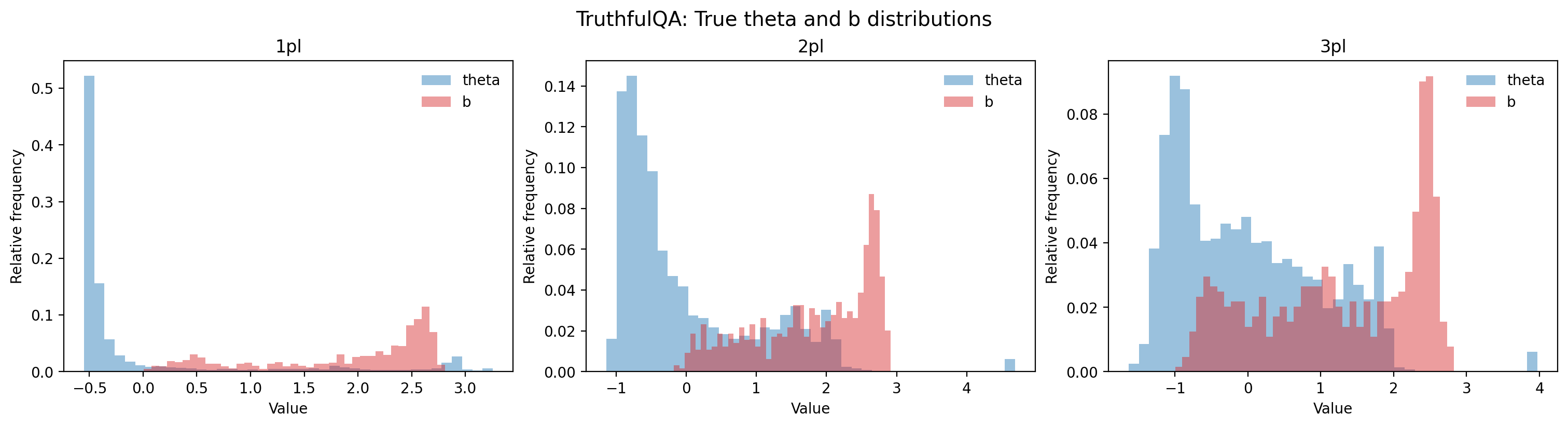}
    \caption{Distributions of true capability ($\theta$) and difficulty ($b$) parameters for \textsc{TruthfulQA}.}
    \label{fig:app_tb_truth}
\end{figure*}

\begin{figure*}[ht]
    \centering
    \includegraphics[width=\linewidth]{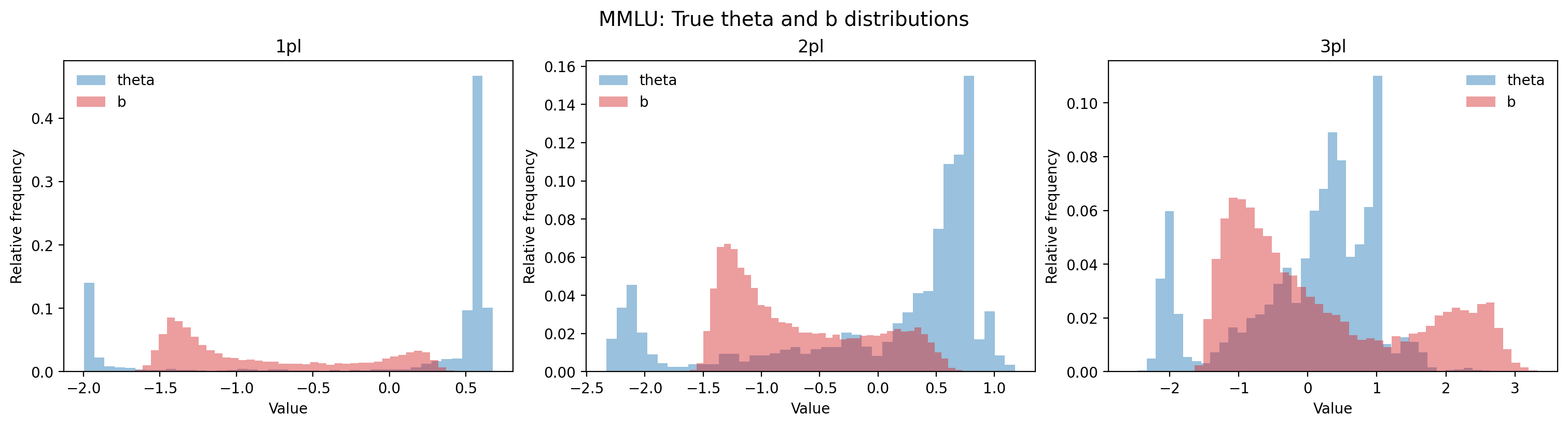}
    \caption{Distributions of true capability ($\theta$) and difficulty ($b$) parameters for \textsc{MMLU}.}
    \label{fig:app_tb_mmlu}
\end{figure*}

In Figs.~\ref{fig:app_tb_truth}--\ref{fig:app_tb_mmlu}, we present distributions of true capability ($\theta$) and difficulty ($b$) parameters for the six benchmarks under 1PL, 2PL, and 3PL models, estimated from the full OpenLLM response matrices via variational inference. All distributions exhibit non-normality, with capability values showing multimodality across all three IRT models.

\begin{figure*}[ht]
    \centering
    \includegraphics[width=\linewidth]{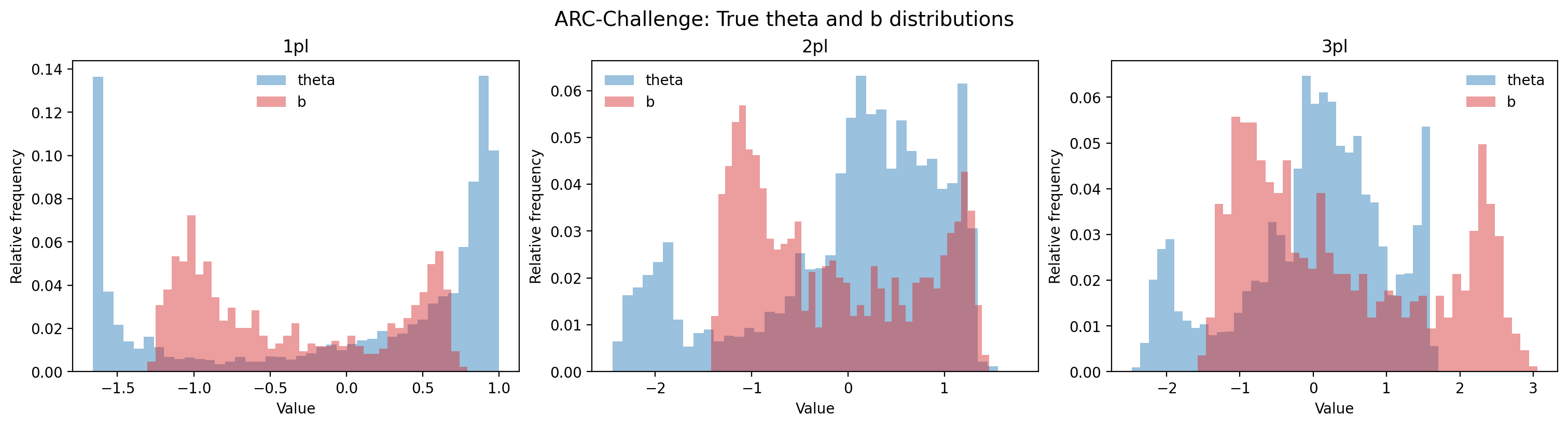}
    \caption{Distributions of true capability ($\theta$) and difficulty ($b$) parameters for \textsc{ARC-Challenge}.}
\end{figure*}

\begin{figure*}[ht]
    \centering
    \includegraphics[width=\linewidth]{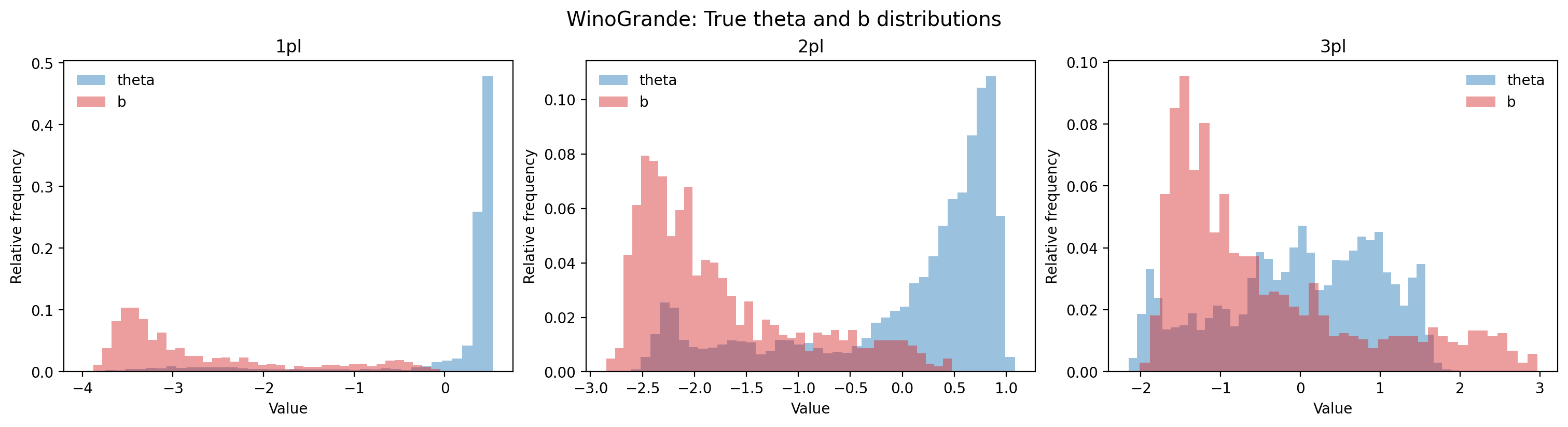}
    \caption{Distributions of true capability ($\theta$) and difficulty ($b$) parameters for \textsc{WinoGrande}.}
\end{figure*}

\begin{figure*}[ht]
    \centering
    \includegraphics[width=\linewidth]{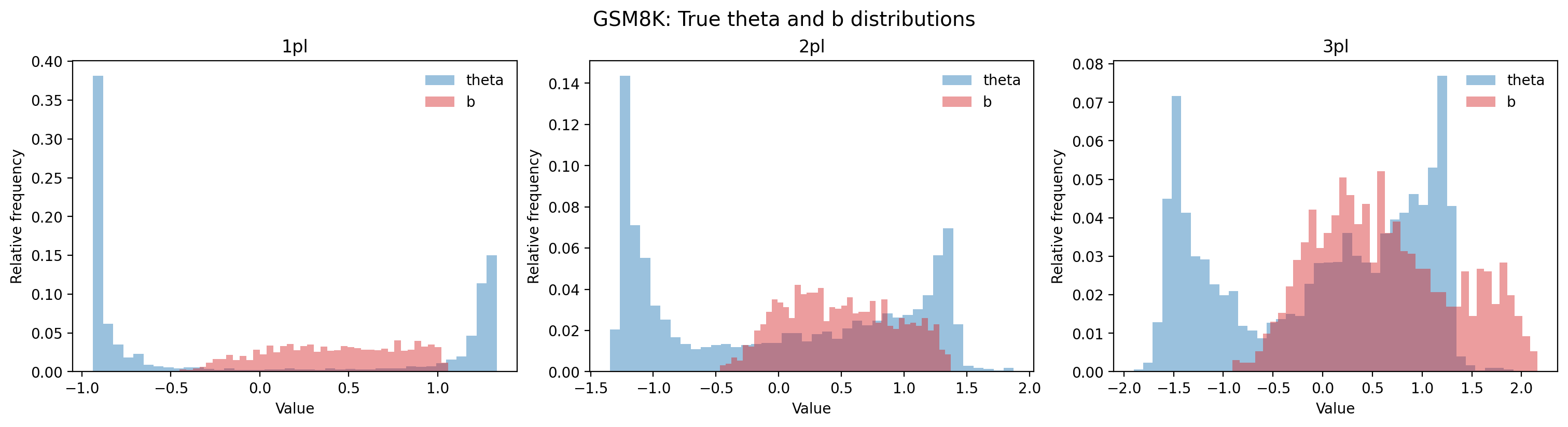}
    \caption{Distributions of true capability ($\theta$) and difficulty ($b$) parameters for \textsc{GSM8K}.}
\end{figure*}

\begin{figure*}[ht]
    \centering
    \includegraphics[width=\linewidth]{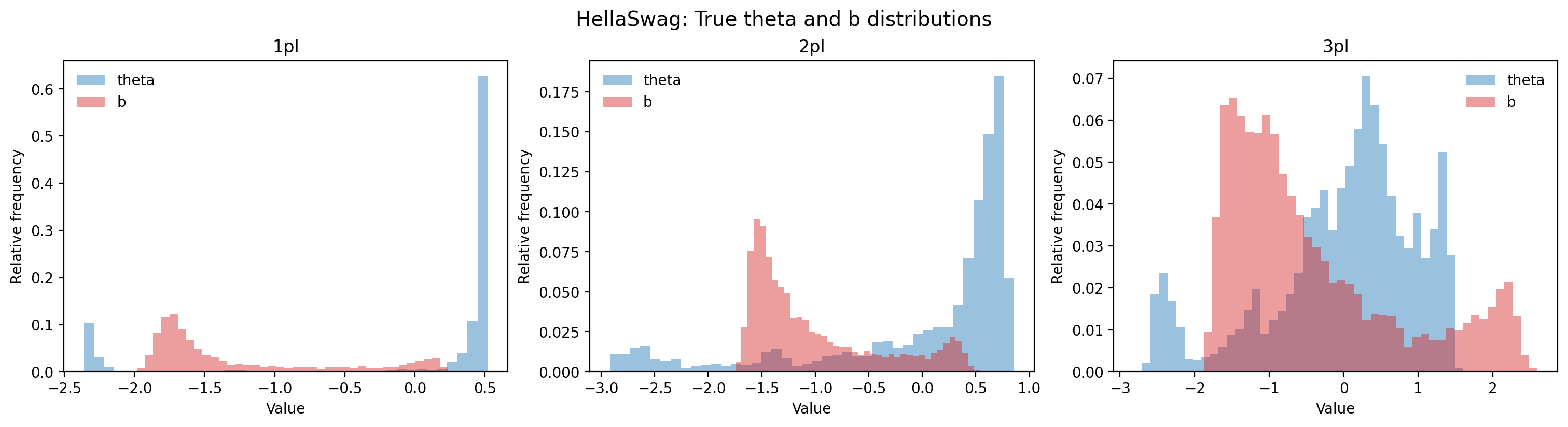}
    \caption{Distributions of true capability ($\theta$) and difficulty ($b$) parameters for \textsc{HellaSwag}.}
\end{figure*}

\begin{figure*}[ht]
    \centering
    \includegraphics[width=\linewidth]{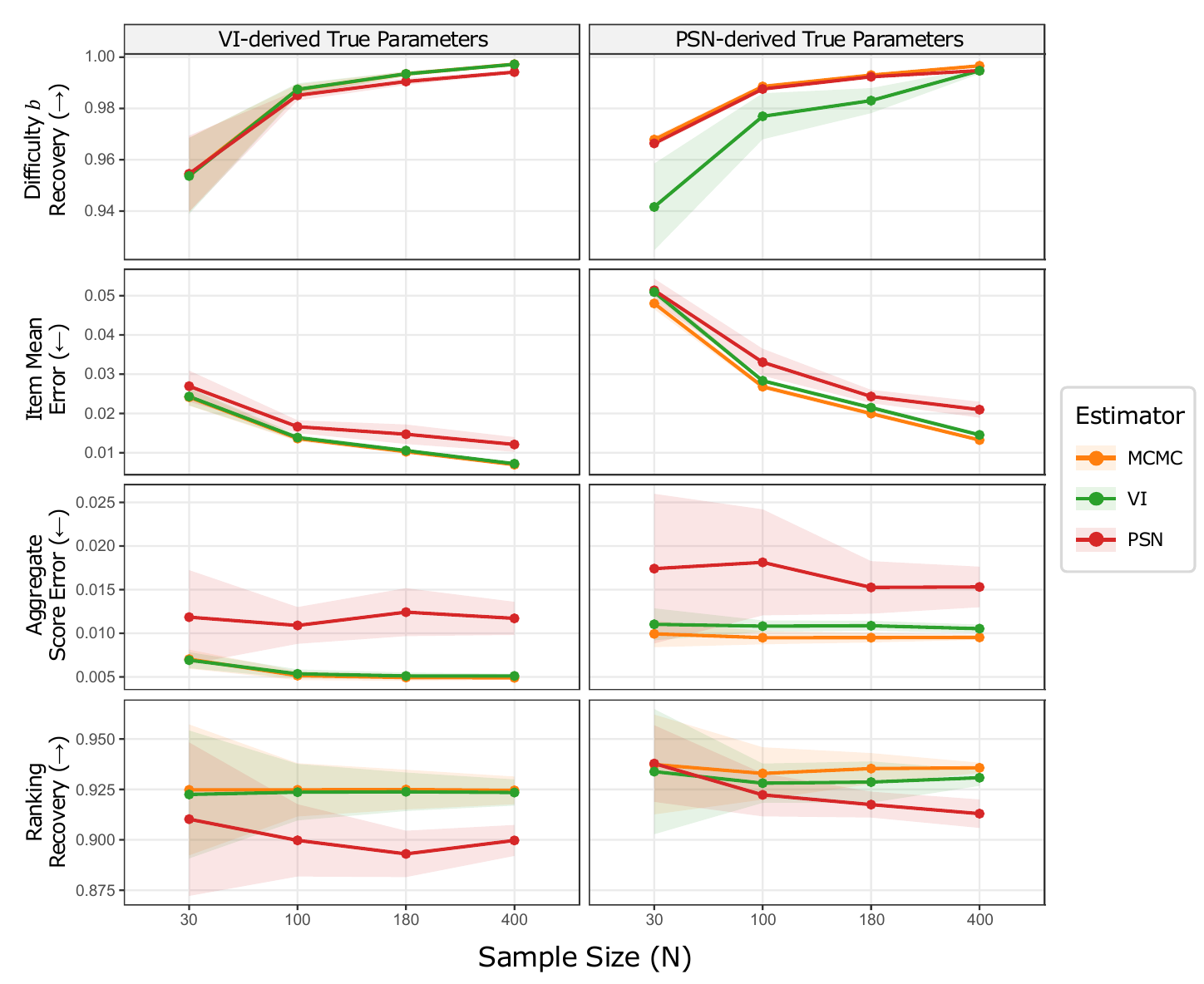}
    \caption{Performance of MCMC, VI, and PSN under VI- and PSN-derived data-generating parameters for ARC-Challenge with 1PL.}
    \label{fig:app_psn_gt}
\end{figure*}

\begin{table}[t]
    \centering
    \small
    \caption{VI Performance under VI- and PSN-derived data-generating parameters for ARC-Challenge with 1PL. 
    Convergence rates are based on 50 replications per sample size under VI-derived parameters and 10 under PSN-derived parameters. 
    Metric values are means with standard deviations across converged replications in parentheses. 
    }
    \label{tab:app_psn_gt}
    \begin{tabular}{llcccc}
        \toprule
        & & \multicolumn{4}{c}{\textbf{Sample Size}} \\
        \cmidrule{3-6}
        \textbf{Metric} & \textbf{True Params.} & $N\!=$ \textit{30} & $N\!=$ \textit{100} & $N\!=$ \textit{180} & $N\!=$ \textit{400} \\
        \midrule
        \multirow{2}{*}{Convergence (\%)} & \textit{VI-derived} & 96 & 98 & 98 & 100 \\
        & \textit{PSN-derived} & 90 & 90 & 90 & 100 \\
        \cmidrule{2-6}
        \multirow{2}{*}{Difficulty Recovery ($\uparrow$)} & \textit{VI-derived} & .954 (.015) & .987 (.002) & .993 (.001) & .997 (.0003) \\
        & \textit{PSN-derived} & .942 (.017) & .977 (.009) & .983 (.005) & .995 (.0012) \\
        \cmidrule{2-6}
        \multirow{2}{*}{Item Mean Error ($\downarrow$)} & \textit{VI-derived} & .0243 (.0023) & .0139 (.0007) & .0106 (.0005) & .0072 (.0002) \\
        & \textit{PSN-derived} & .0509 (.0012) & .0283 (.0005) & .0215 (.0006) & .0146 (.0005) \\
        \cmidrule{2-6}
        \multirow{2}{*}{Aggregate Score Error ($\downarrow$)} & \textit{VI-derived} & .0069 (.0009) & .0053 (.0005) & .0051 (.0004) & .0051 (.0002) \\
        & \textit{PSN-derived} & .0110 (.0018) & .0108 (.0007) & .0109 (.0006) & .0105 (.0005) \\
        \cmidrule{2-6}
        \multirow{2}{*}{Ranking Recovery ($\uparrow$)} & \textit{VI-derived} & .923 (.032) & .924 (.014) & .924 (.010) & .923 (.006) \\
        & \textit{PSN-derived} & .934 (.031) & .928 (.010) & .929 (.010) & .931 (.004) \\
        \bottomrule
    \end{tabular}
\end{table}

\subsection{Reliability of the data-generating parameters}\label{app:true_params}

Our primary simulations introduced in Sec.~\ref{sec:4} used parameters estimated by VI from the empirical OpenLLM response matrices as data-generating parameters. 
These estimates should be understood as pseudo-ground truth rather than verified true parameters. 
Their suitability raises two related questions: 

\textit{Whether VI estimation errors could make the simulated data unrealistic.}
The strongest VI instability in our simulations occurred at small sample sizes, whereas each empirical response matrix used to derive the data-generating parameters contained more than 5,000 models. 
This is well beyond the small-sample regime in which VI exhibited poor parameter recovery, providing some support for the stability of the data-generating parameters. 
Exact capability rankings remain a caveat, because irregular capability distributions can make latent ordering difficult to recover even with many models. 
Nevertheless, the overall shape of the capability distribution is less sensitive than the ordering of closely matched models and was broadly consistent with the distribution of observed aggregate scores. 
Thus, the empirical fits provide a plausible basis for simulation, although they should not be interpreted as known ground truth.

\textit{Whether using VI both to generate the parameters and as a competing estimator could favor VI in the subsequent recovery analysis.}
After fitting VI, we transformed the parameters so that the data-generating capabilities had mean 0 and standard deviation 1, with the corresponding transformations applied to the item parameters. 
This removes arbitrary location and scale differences across fitted models. 
However, rescaling alone cannot rule out estimator-specific bias arising from VI's dual role in parameter generation and evaluation. 
Notably, VI did not consistently outperform the other estimators in our primary experiments and often showed poorer item-parameter recovery at small sample sizes, providing no evidence of a general advantage from this dual role.

Therefore, to further assess this possibility, we conducted a sensitivity analysis on \textsc{ARC-Challenge} under 1PL, comparing MCMC, VI, and PSN at $N\in{30,100,180,400}$.  
We fitted PSN to the same empirical response matrix and generated 10 replications per condition using the resulting parameters. 
We then evaluated difficulty recovery, item mean error, aggregate score error, and ranking recovery, comparing the relative performance of the three estimators under PSN- and VI-derived data-generating parameters.

As shown in Fig.~\ref{fig:app_psn_gt}, changing the source of the data-generating parameters did not substantially alter the relative performance of MCMC, VI, and PSN across most metrics. 
Difficulty recovery was the main exception, with PSN performing more favorably relative to VI under PSN-derived parameters. 
This exception does not reverse our broader finding in Sec.~\ref{sec:5.3} that VI can exhibit unreliable item-parameter recovery at small sample sizes. 
In particular, VI did not gain a consistent advantage when evaluated under VI-derived parameters across the outcomes.
Table~\ref{tab:app_psn_gt} reports the absolute performance of VI under the two data-generating parameter sources. 

Overall, this sensitivity analysis suggests that our main comparisons are not driven solely by the use of VI-derived data-generating parameters. 

\section{Exit statuses of MML-EM and MCMC}\label{app:exit}
\begin{figure*}[ht]
    \centering
    \includegraphics[width=\linewidth]{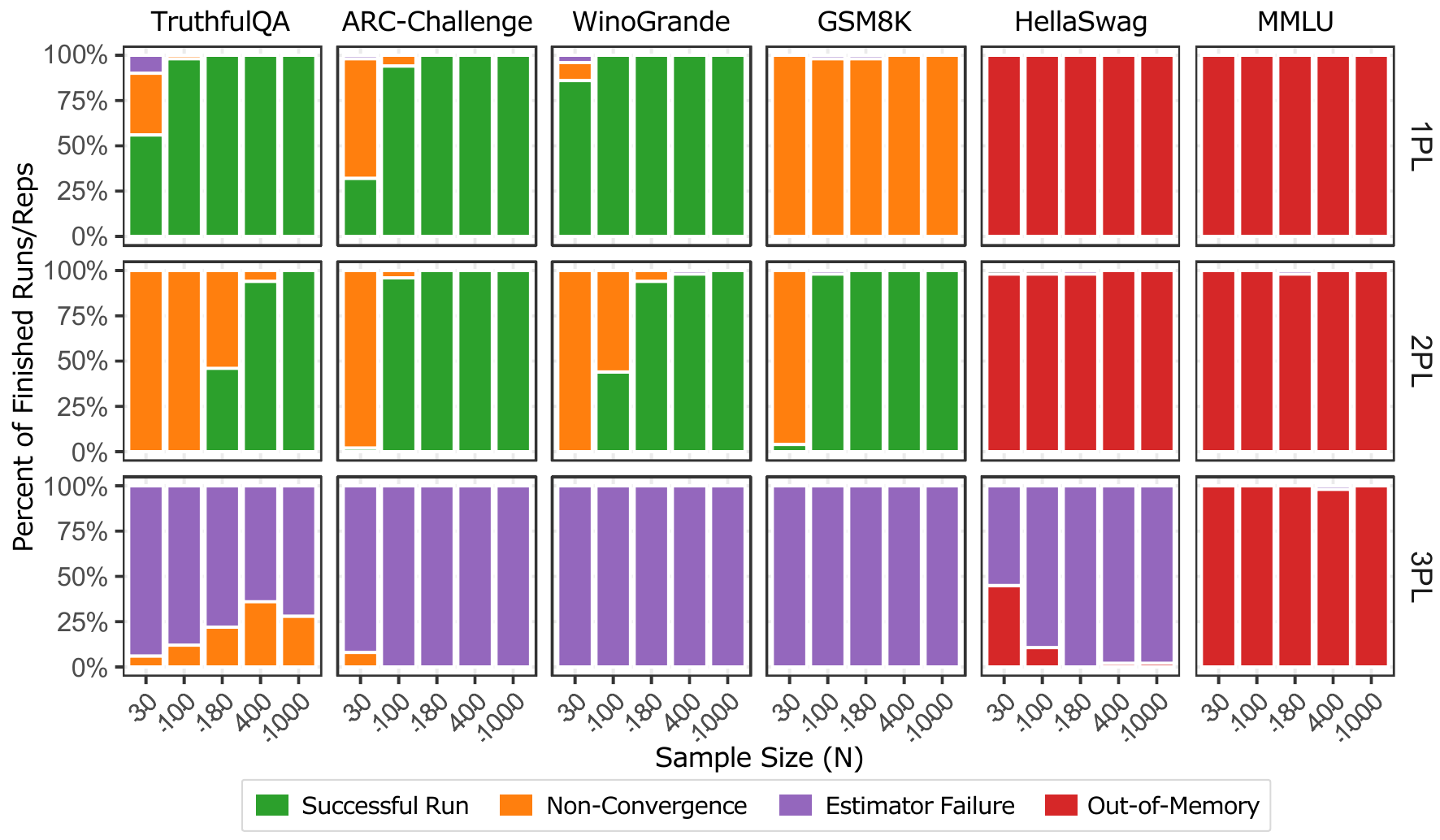}
    \caption{Distribution of MML-EM exit statuses across IRT models, benchmarks, and sample sizes. \textsc{HellaSwag} and \textsc{MMLU} failed entirely due to out-of-memory errors. For smaller benchmarks, non-convergence was the primary failure mode, particularly at small sample sizes and under 3PL. Success rates improved with increasing sample size for 1PL and 2PL but remained near zero for 3PL across all conditions.}
    \label{fig:app_exit_em}
\end{figure*}

\begin{figure*}[ht]
    \centering
    \includegraphics[width=\linewidth]{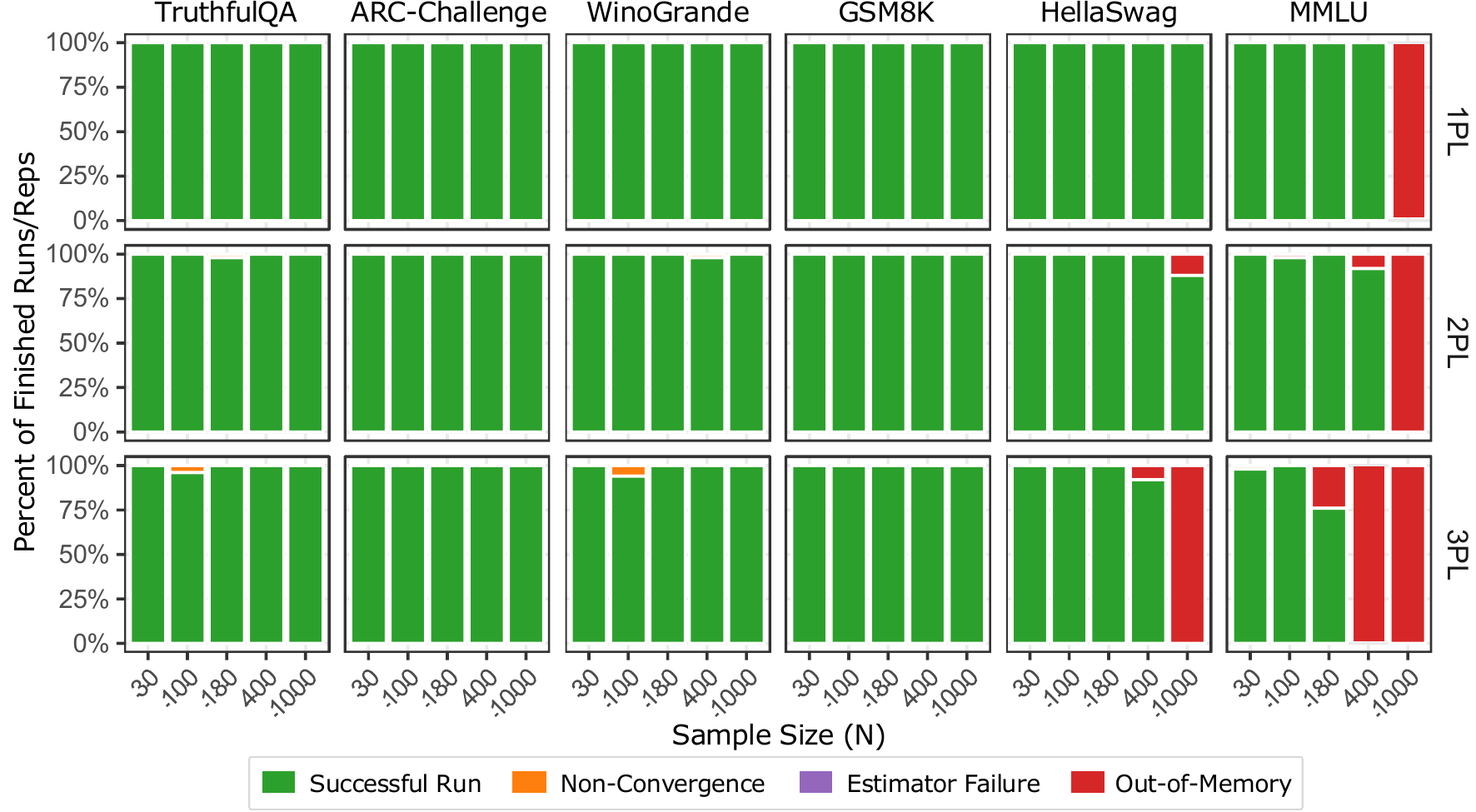}
    \caption{Distribution of MCMC exit statuses across IRT models, benchmarks, and sample sizes.}
    \label{fig:app_exit_mcmc}
\end{figure*}

The distributions of the exit statuses of MML-EM and MCMC are displayed in Figs.~\ref{fig:app_exit_em} \&~\ref{fig:app_exit_mcmc}.

\begin{figure*}[ht]
    \centering
    \includegraphics[width=\linewidth]{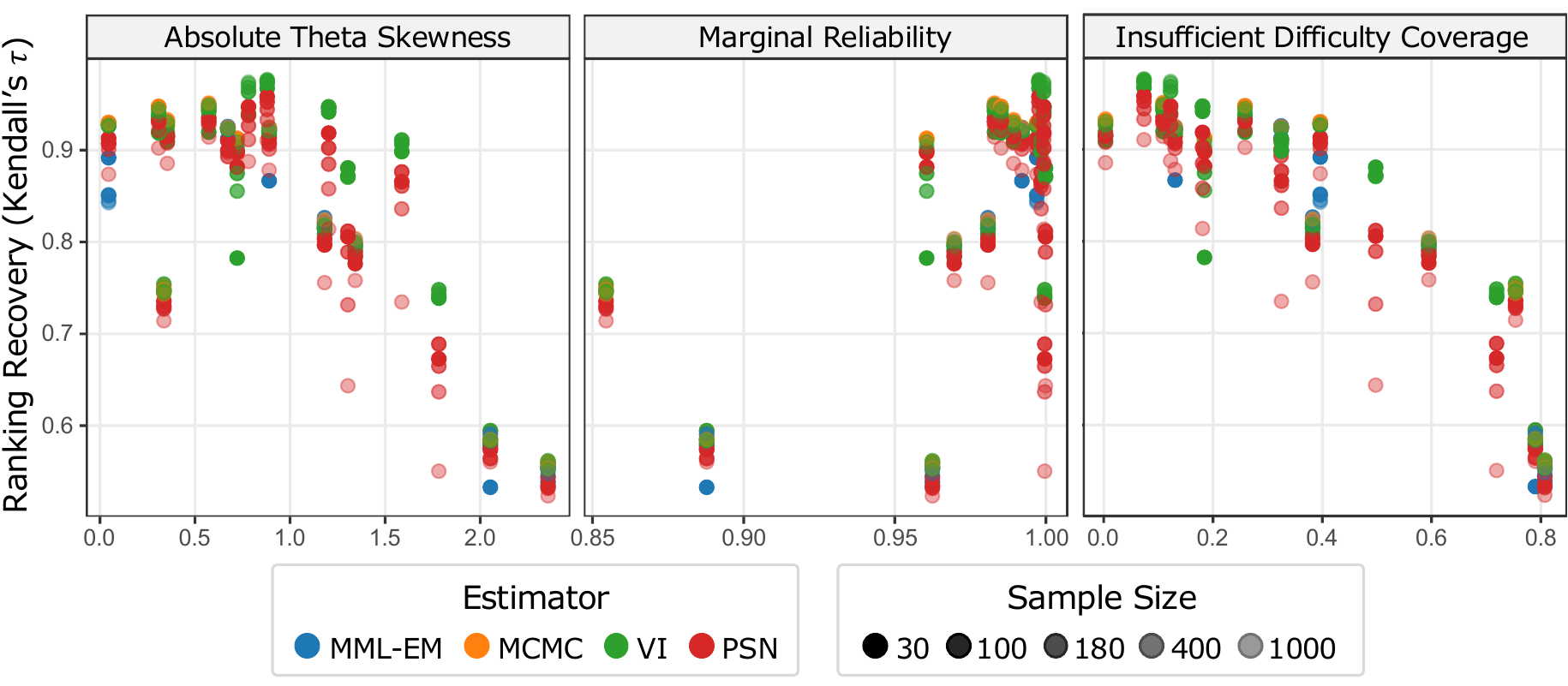}
    \caption{Bivariate relationships between ranking recovery and absolute theta skewness, marginal reliability, and insufficient difficulty coverage, defined in App.~\ref{app:failure_analysis}. 
    Each point represents a simulation condition; color indicates the estimator, and opacity indicates sample size. 
    }
    \label{fig:app_ranking_factors}
\end{figure*}

\begin{table}[ht]
    \centering
    \caption{Multilevel regression predicting Fisher-transformed ranking recovery. 
    Continuous predictors were standardized, while estimator was effect-coded such that each estimator coefficient represents its deviation from the mean across all four estimators. 
    Bold and italics indicate the largest and second-largest absolute coefficients among the continuous predictors, respectively.
    }
    \label{tab:app_factors_1}
    \begin{tabular}{lcc}
        \toprule
        \textbf{Predictor} &  \textbf{Coefficient} & \textbf{95\% Jackknife CI} \\
        \midrule
        Intercept & 1.380           & [1.297, 1.462] \\
        \midrule
        Absolute Theta Skewness & \textit{-0.103} & [-0.181, -0.026] \\
        Marginal Reliability & 0.064           & [-0.020, 0.149] \\
        Insufficient Difficulty Coverage & \textbf{-0.280} & [-0.407, -0.152] \\
        Log Sample Size & -0.026          & [-0.041, -0.011] \\
        Estimator (MML-EM) & -0.019          & [-0.110, 0.071] \\
        Estimator (MCMC) & 0.082           & [0.040, 0.124] \\
        Estimator (PSN) & -0.111          & [-0.166, -0.057] \\
        Estimator (VI) & 0.049           & [0.006, 0.091] \\
        \bottomrule
    \end{tabular}
\end{table}

\begin{table}[ht]
    \centering
    \caption{Marginal $R^2$ values for models predicting Fisher-transformed ranking recovery. 
    The base model included estimator and log sample size, and each extended model added one standardized predictor in Table~\ref{tab:app_factors_1}. 
    Marginal $R^2$ represents the variance explained by the fixed effects. 
    Bold indicates the model with the highest marginal $R^2$.}
    \label{tab:app_factors_2}
    \begin{tabular}{lc}
    \toprule
    \textbf{Model} & \textbf{Marginal} $R^2$\\
    \midrule
    Base (\textit{Estimator} + \textit{Log Sample Size}) & .039\\
    Base + \textit{Absolute Theta Skewness} & .437\\
    Base + \textit{Marginal Reliability} & .325\\
    Base + \textit{Insufficient Difficulty Coverage} & \textbf{.760}\\
    \bottomrule
    \end{tabular}
\end{table}

\section{Failure mode analysis for ranking recovery}
\label{app:failure_analysis}

To examine the factors associated with the poor ranking recovery in Sec.~\ref{sec:5.2}, we first fitted a multilevel linear regression model predicting the Fisher-transformed ranking-recovery correlation. 
All continuous predictors were standardized, such that their coefficients represent the expected change in the outcome associated with a one-standard-deviation increase in the predictor. 
The model included the following predictors:
\begin{itemize}
    \item \textit{Absolute skewness} $|\gamma|$, defined in Sec.~\ref{sec:5.2} as the absolute sample skewness of the true model capability $\theta$;
    \item \textit{Insufficient difficulty coverage}, defined as the proportion of models having $\le$ 5 items whose difficulty is in the $\theta \pm .2$ neighborhood;
    \item \textit{Marginal reliability}~\citep{bechger2003using}, with lower values indicating less precise capability measurement due to limited test information, which may reflect difficulty mismatch, low item discrimination, or both;
    \item \textit{Log sample size}, defined as $\log(N)$; and
    \item \textit{Estimator}, effect-coded such that each estimator's coefficient reflects its deviation from the mean across all four estimators introduced in Sec.~\ref{sec:4.2}.
\end{itemize}

Table~\ref{tab:app_factors_1} reports the regression coefficients. 
For the four continuous predictors, a larger absolute coefficient indicates a stronger conditional association with ranking recovery per standard-deviation change, holding the other variables fixed. 
The estimator coefficients instead represent effect-coded deviations from the mean across estimators. 
This regression model accounted for approximately 80\% of variance in ranking recovery ($R^2=.798$).

Fig.~\ref{fig:app_ranking_factors} presents bivariate scatter plots for the three data-characteristic predictors with larger coefficients.
We additionally assessed their incremental predictive value by fitting three regression models that separately added each predictor to a base model containing only estimator and log sample size, and reporting the resulting marginal $R^2$ in Table~\ref{tab:app_factors_2}.

Together, these analyses helped clarify the factors associated with ranking failure.
insufficient difficulty coverage was the most predictive factor of poor ranking recovery: adding it to the base model increased the marginal $R^2$ from $.039$ to $.760$. 
Absolute skewness was the second most predictive. 
It was moderately correlated with insufficient difficulty coverage (Kendall's correlation $=.40$) but also increased incremental predictive power.
On the estimation side, larger sample size did not improve ranking recovery and instead showed a small negative association, consistent with Fig.~\ref{fig:skewness_ranking}. 
Relative to the mean across estimators, MCMC and VI achieved better ranking recovery, PSN performed worse, and MML-EM showed no clear difference.

\section{Detailed results for parameter recovery}\label{app:recovery}

\subsection{Aggregate score error}\label{app:capability_recovery}
\begin{figure*}[ht]
    \centering
    \includegraphics[width=\linewidth]{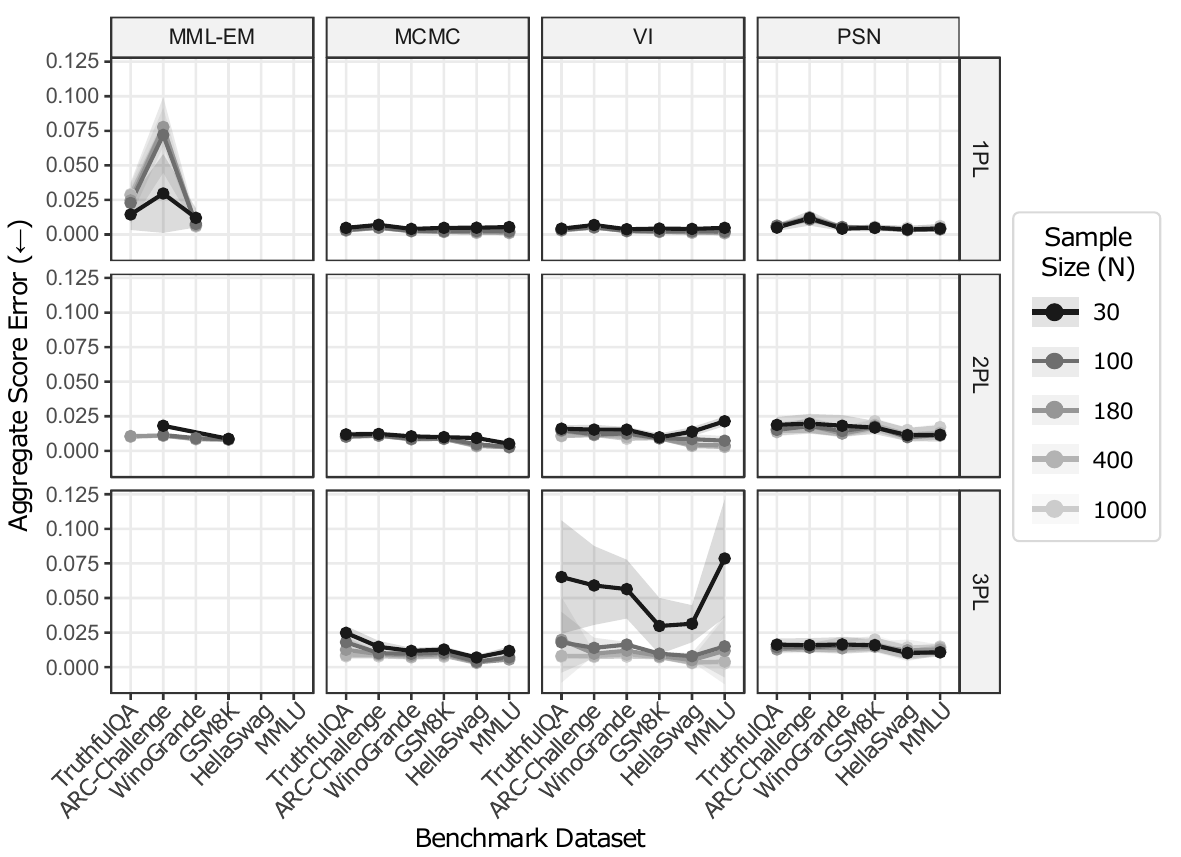}
    \caption{Aggregate score error across IRT models, estimators, benchmarks, and sample sizes. 
    Lower is better. 
    Shaded regions indicate $\pm$1 standard deviation across 50 replications. 
    Missing points indicate computational infeasibility.}
    \label{fig:app_agg_err}
\end{figure*}

Please refer to Fig.~\ref{fig:app_agg_err} for detailed results. 

\subsection{Item parameter recovery}\label{app:item_recovery}
\begin{figure*}[ht]
    \centering
    \includegraphics[width=\linewidth]{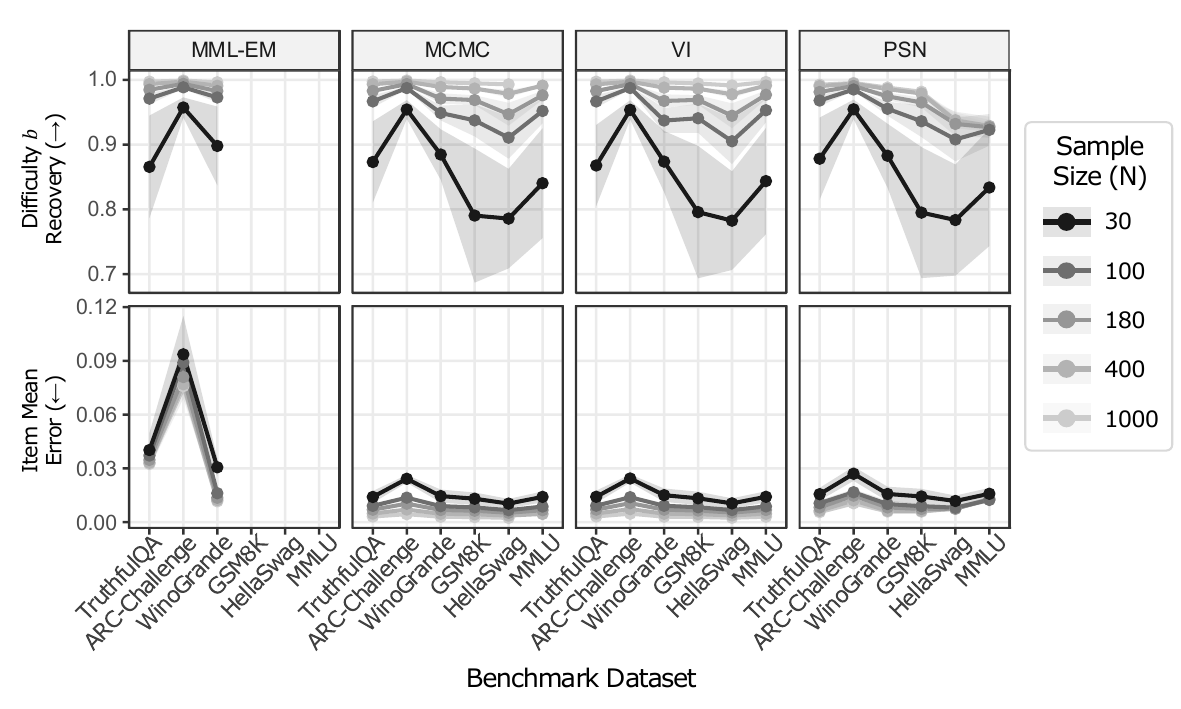}
    \caption{Item parameter recovery under 1PL across estimators, benchmarks, and sample sizes. 
    Higher is better for difficulty recovery, and lower is better for item mean error. 
    Shaded regions indicate $\pm$1 standard deviation across 50 replications. 
    Missing points indicate computational infeasibility.}
    \label{fig:app_1pl_item}
\end{figure*}

\begin{figure*}[ht]
    \centering
    \includegraphics[width=\linewidth]{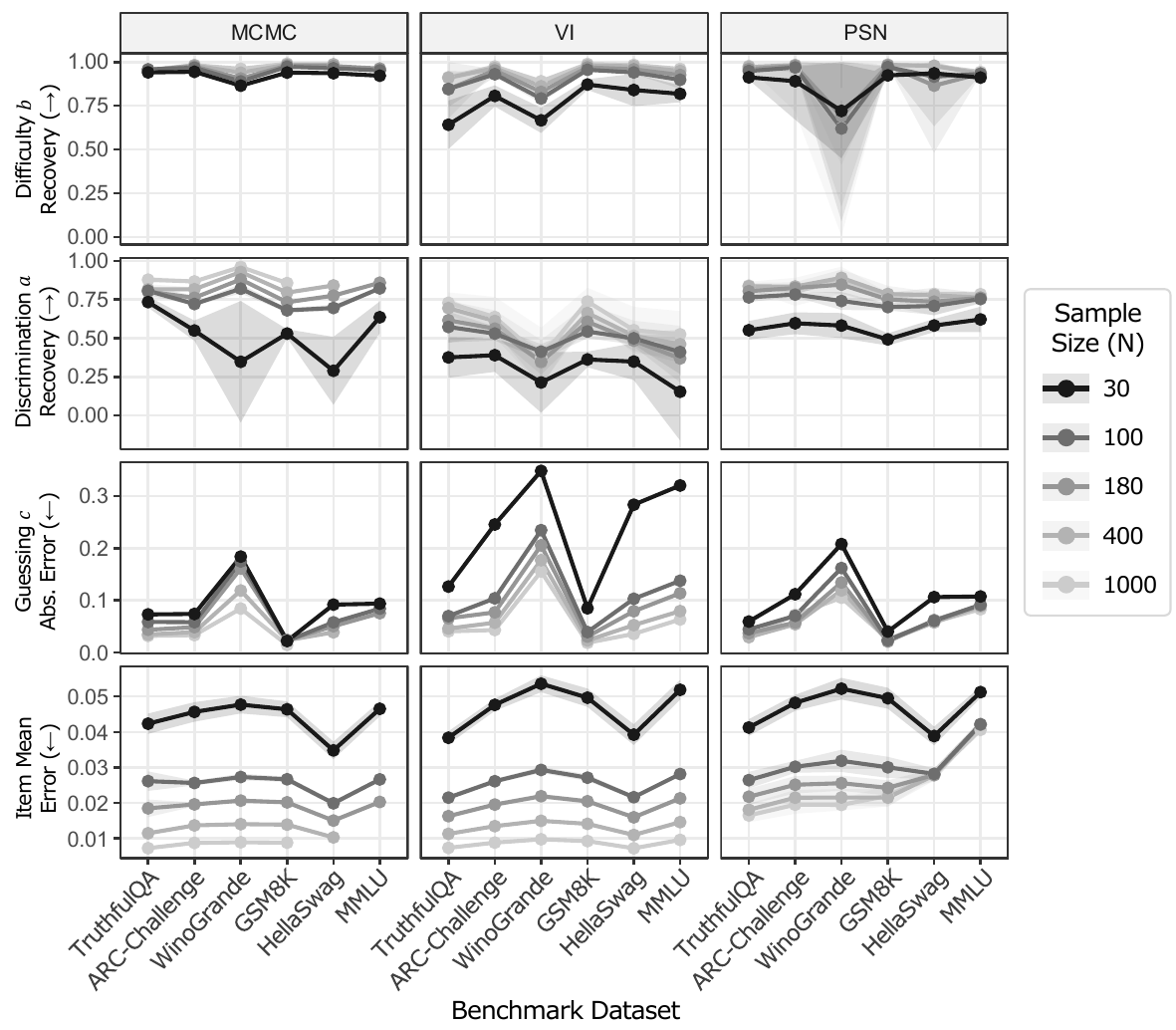}
    \caption{Item parameter recovery under 3PL across estimators, benchmarks, and sample sizes. 
    Higher is better for difficulty recovery and discrimination recovery; lower is better for guessing absolute error and item mean error. 
    Shaded regions indicate $\pm$1 standard deviation across 50 replications. 
    Missing subplots and points indicate computational infeasibility.}
    \label{fig:app_3pl_item}
\end{figure*}

Please refer to Figs.~\ref{fig:app_1pl_item}~\&~\ref{fig:app_3pl_item} for detailed results. 

\begin{figure*}[ht]
    \centering
    \includegraphics[width=\linewidth]{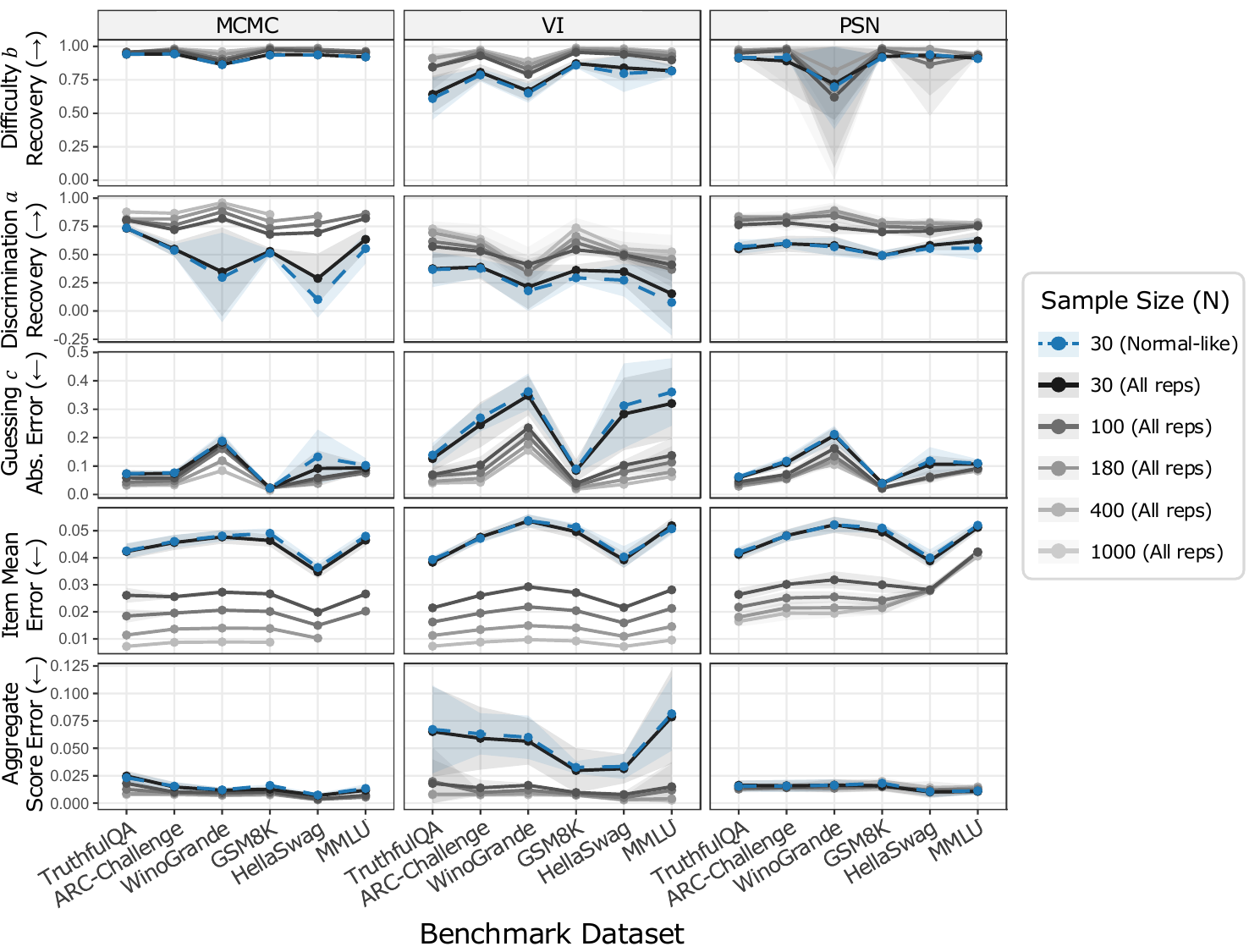}
    \caption{Parameter and aggregate-score recovery under 3PL for the near-normal $N=30$ subset (blue), compared with all replications at each sample size (grayscale). 
    Points and shaded bands show the mean and one standard deviation across replications, respectively. }
    \label{fig:app_post_hoc_1}
\end{figure*}

\begin{figure*}[ht]
    \centering
    \includegraphics[width=\linewidth]{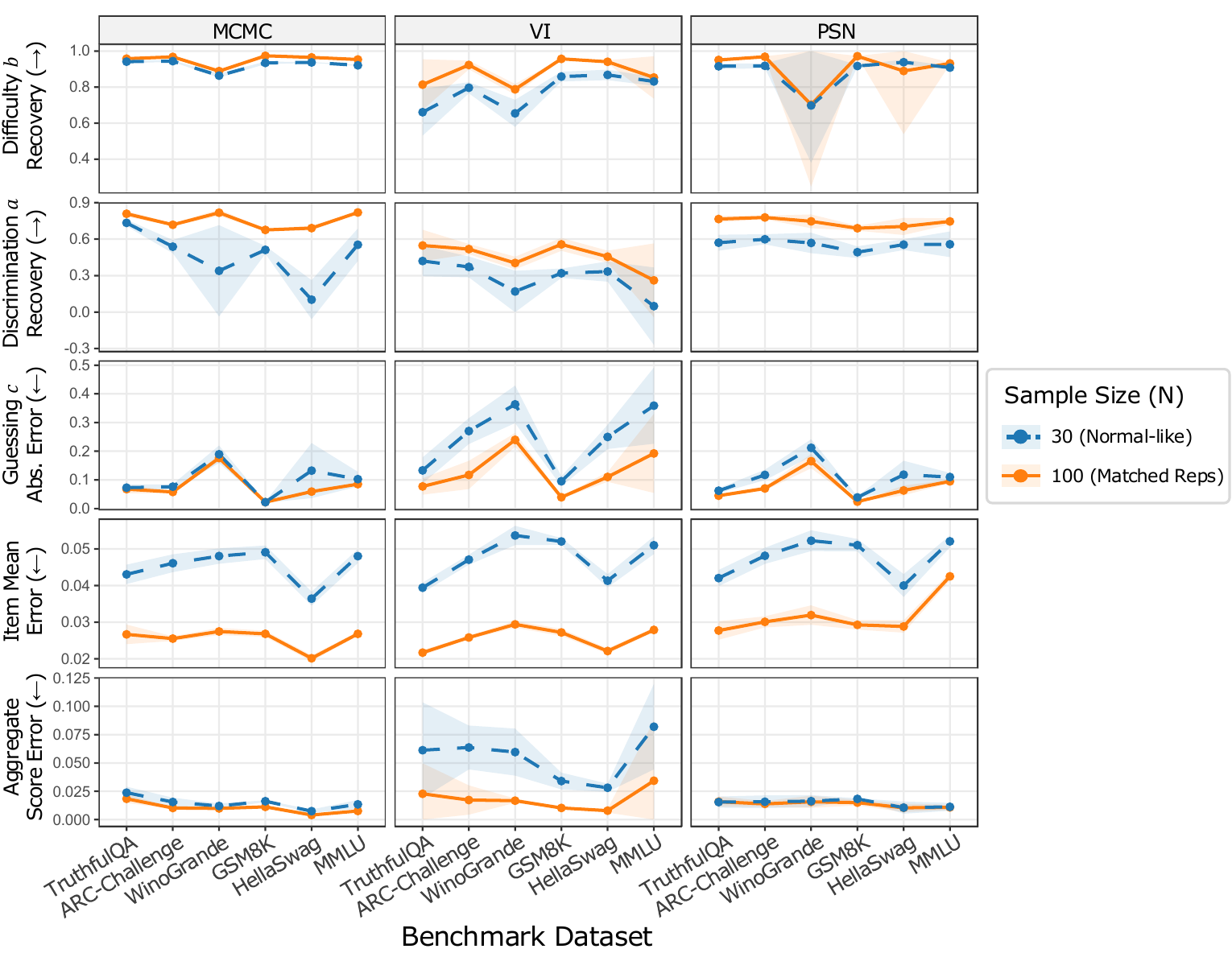}
    \caption{Parameter and aggregate-score recovery under 3PL for the near-normal $N=30$ subset (blue), compared with the matched $N=100$ results from the same selected replications (orange).
    Points and shaded bands show the mean and one standard deviation across replications, respectively. }
    \label{fig:app_post_hoc_2}
\end{figure*}

\begin{figure*}[ht]
    \centering
    \includegraphics[width=\linewidth]{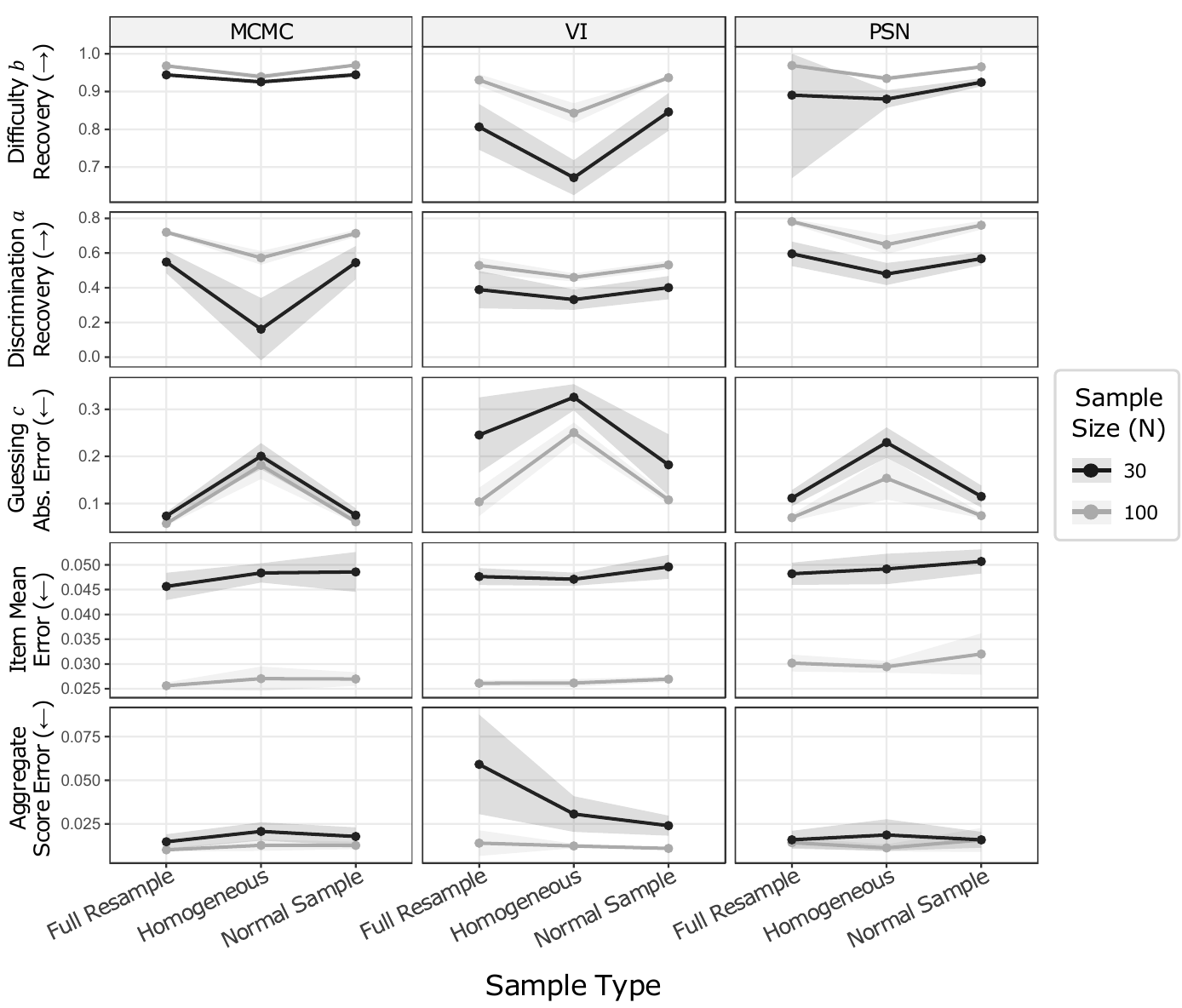}
    \caption{Parameter and aggregate-score recovery for ARC-Challenge under 3PL at $N=30$ and $N=100$ across three capability-sampling types: \textit{full resample} draws from the complete empirical capability distribution, \textit{homogeneous} resample from its $[-1,1]$ region, and \textit{normal sample} from the standard normal distribution. 
    Points and shaded bands show the mean and one standard deviation across replications, respectively.}
    \label{fig:app_normal}
\end{figure*}

\subsection{Disentangling sample size from distributional skewness}\label{app:disentangle}

As discussed in Sec.~\ref{sec:2.1}, the empirical capability distributions observed in AI evaluation context are frequently skewed, multimodal, or otherwise irregular. 
Because these properties may themselves impair IRT estimation, the poor parameter recovery observed at $N=30$ could reflect both insufficient sample size and capability-distribution shape. 
The large item sets common in AI benchmarks can improve capability estimation because each model capability is informed by many item responses. 
However, they do not increase the number of observations available for each item parameter, which remains informed by only $N$ model responses. 
Thus, a large item set cannot necessarily compensate for a small model sample in item-parameter recovery.

We therefore conducted two complementary analyses to assess whether the recovery improvement from $N=30$ to $N=100$ persisted under more regular capability distributions.

\textit{Post-hoc analysis of near-normal replications}.
At $N=30$, we retained replications whose distance did not exceed the 95th percentile of the distances obtained from standard-normal samples of the same size, as measured by the 2-Wasserstein distance.
This criterion identifies empirical samples whose departures from normality are comparable to those arising from finite random samples drawn from a normal distribution.

Among the $6\times3\times50=900$ existing $N=30$ replications, 119 met this criterion. Of these, 112 were from the 3PL simulations: 36 for \textsc{WinoGrande}, 27 for \textsc{ARC-Challenge}, 17 each for \textsc{TruthfulQA} and \textsc{HellaSwag}, 12 for \textsc{MMLU}, and 3 for \textsc{GSM8K}; we focused on these 112 replications. 
For each benchmark $\times$ estimator condition, we recomputed the mean and standard deviation of item parameter recovery, item mean error, and aggregate score error.

Fig.~\ref{fig:app_post_hoc_1} compares this near-normal $N=30$ subset with the complete results. 
Across metrics, the near-normal subset generally tracked the full $N=30$ sample and remained substantially worse than the $N=100$ results. 
Fig.~\ref{fig:app_post_hoc_2} provides a matched comparison between the selected $N=30$ replications and their corresponding $N=100$ replications. 
Recovery still improved substantially at $N=100$, indicating that restricting attention to near-normal capability samples did not close the small-sample recovery gap. 
The \textsc{GSM8K} results should be interpreted cautiously because only three replications met the near-normality criterion.

This post-hoc analysis does not fully isolate capability-distribution shape because the replications were selected rather than experimentally generated. 
Nevertheless, it shows that the observed $N=30$ recovery problems were not confined to the most irregular empirical capability samples.

\textit{Controlled capability-distribution simulations.}
We next conducted controlled simulations for \textsc{ARC-Challenge} under 3PL, comparing MCMC, VI, and PSN at $N\in{30,100}$. We considered three capability-sampling schemes:
\begin{itemize}
\item \textit{Full resample}: capabilities were resampled from the complete empirical capability distribution used in the primary simulations;
\item \textit{Homogeneous resample}: capabilities were resampled from the approximately unimodal $[-1,1]$ region of the empirical distribution, representing a more homogeneous model population; and
\item \textit{Normal sample}: capabilities were drawn from the standard normal distribution, preserving a wider capability range than the homogeneous resample.
\end{itemize}

The full-resample results were based on the original 50 replications, while the homogeneous and normal simulations used 10 replications per sample size. 
Fig.~\ref{fig:app_normal} reports difficulty, discrimination, and guessing-parameter recovery, mean item error, and aggregate score error under the three sampling schemes.

Across both the homogeneous and normal capability distributions, the recovery gap between $N=30$ and $N=100$ persisted for most estimators and metrics. 
Item-parameter recovery generally improved and mean item error declined at $N=100$, consistent with the results under the full empirical distribution. 
VI aggregate score error was the only clear exception to this pattern. 
Thus, replacing the irregular empirical capability distribution with a standard normal distribution or a more homogeneous empirical subset did not eliminate the small-sample recovery problem.

Both analyses show substantial recovery improvements from $N=30$ to $N=100$ under more regular capability distributions, indicating that the poor item-parameter recovery at $N=30$ cannot be explained solely by skewness or other irregularities in the empirical capability distributions. 
Distributional shape remains relevant, particularly for ranking recovery, but small sample size contributes separately to unreliable item-related inference. 
The results therefore support our recommendation that increasing the number of evaluated models is important for item-parameter and aggregate-score recovery even when the underlying capability distribution is approximately normal or comparatively homogeneous.

\begin{figure*}[ht]
    \centering
    \includegraphics[width=0.9\linewidth]{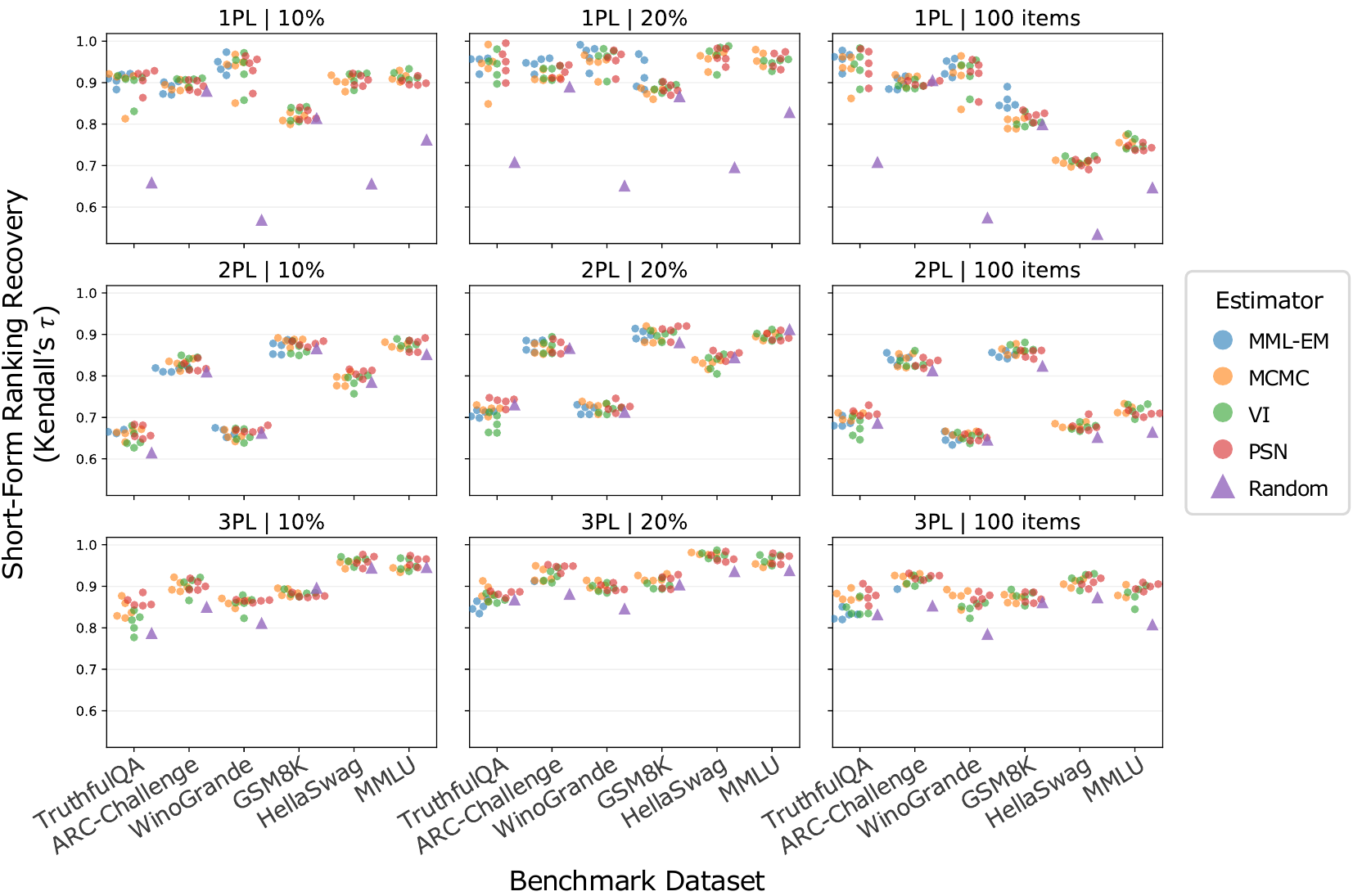}
    \caption{Short-form ranking recovery across test lengths, IRT models, benchmarks, and estimators. 
    Rows correspond to IRT models and columns to subset sizes; sample sizes are pooled without visual distinction.
    Purple triangles show the random baseline averaged over 30 randomly selected subsets of the corresponding size per replication.}
    \label{fig:app_efficient_grid}
\end{figure*}

\begin{figure*}[ht]
    \centering
    \includegraphics[width=\linewidth]{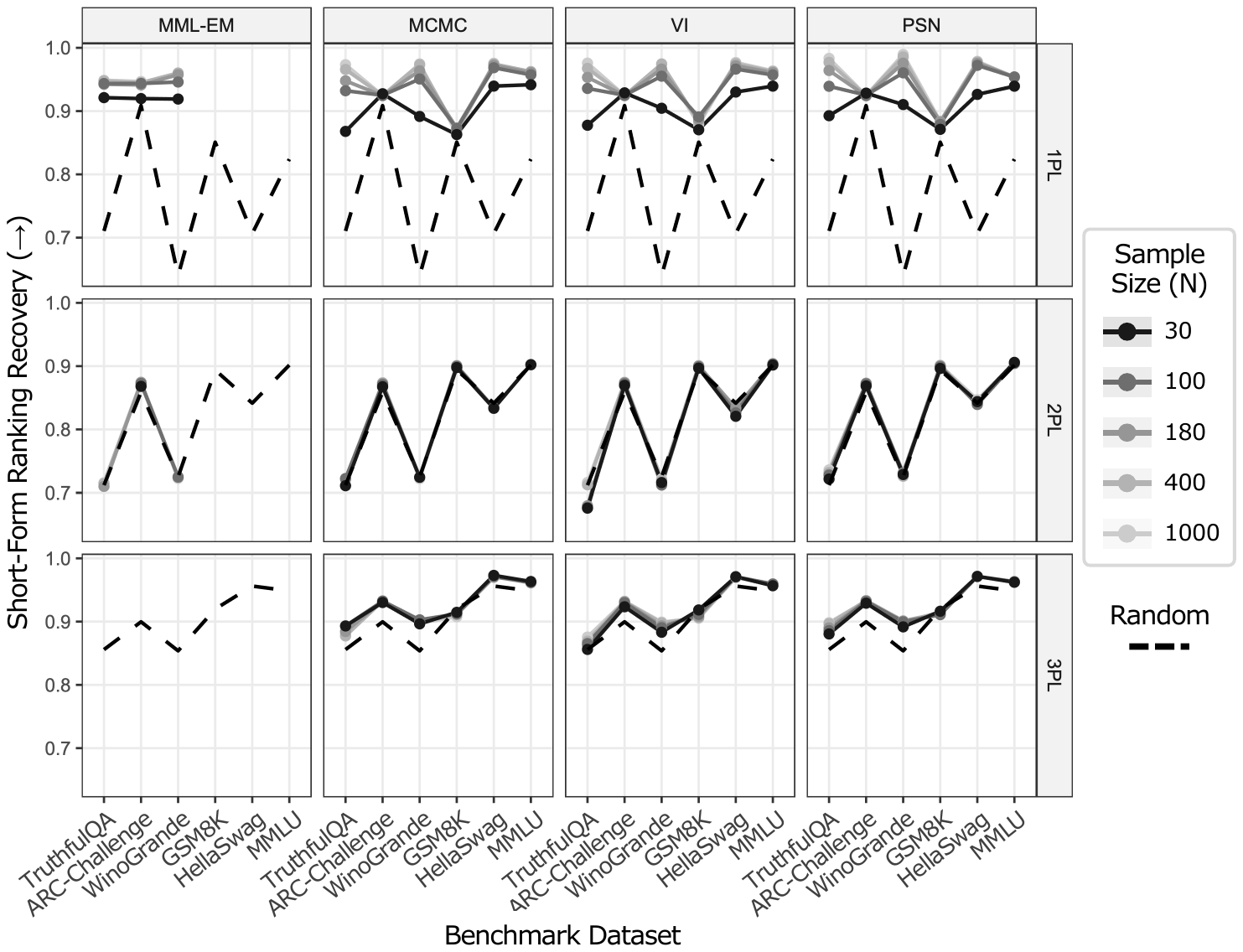}
    \caption{Short-form ranking recovery for 20\%-item subsets across IRT models, benchmarks, estimators, and sample sizes. 
    }
    \label{fig:app_20p}
\end{figure*}

\begin{figure*}[ht]
    \centering
    \includegraphics[width=\linewidth]{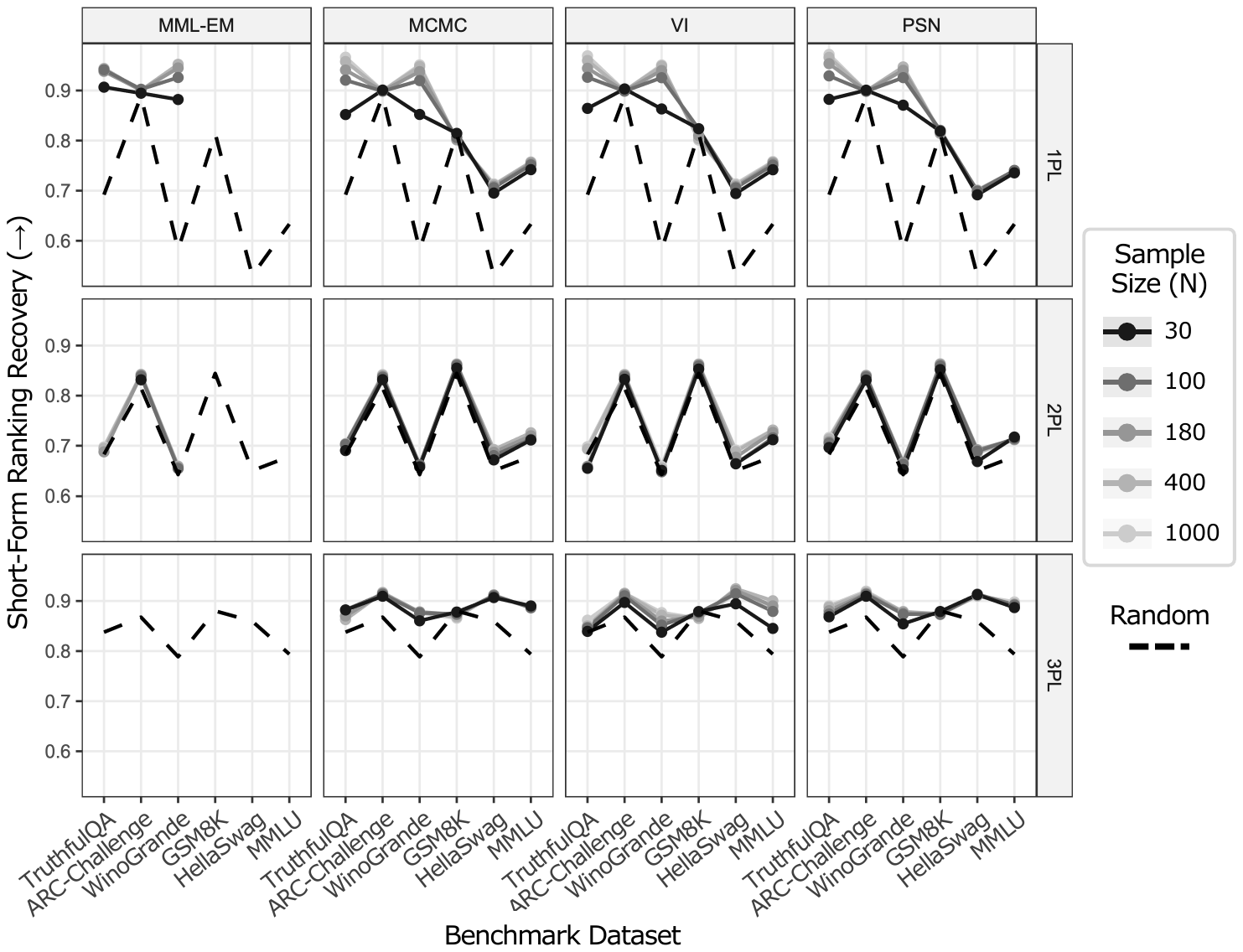}
    \caption{Short-form ranking recovery for 100-item subsets across IRT models, benchmarks, estimators, and sample sizes.}
    \label{fig:app_100}
\end{figure*}

\section{Detailed results for efficient benchmarking}\label{app:short_forms}
\begin{figure*}[ht]
    \centering
    \includegraphics[width=0.9\linewidth]{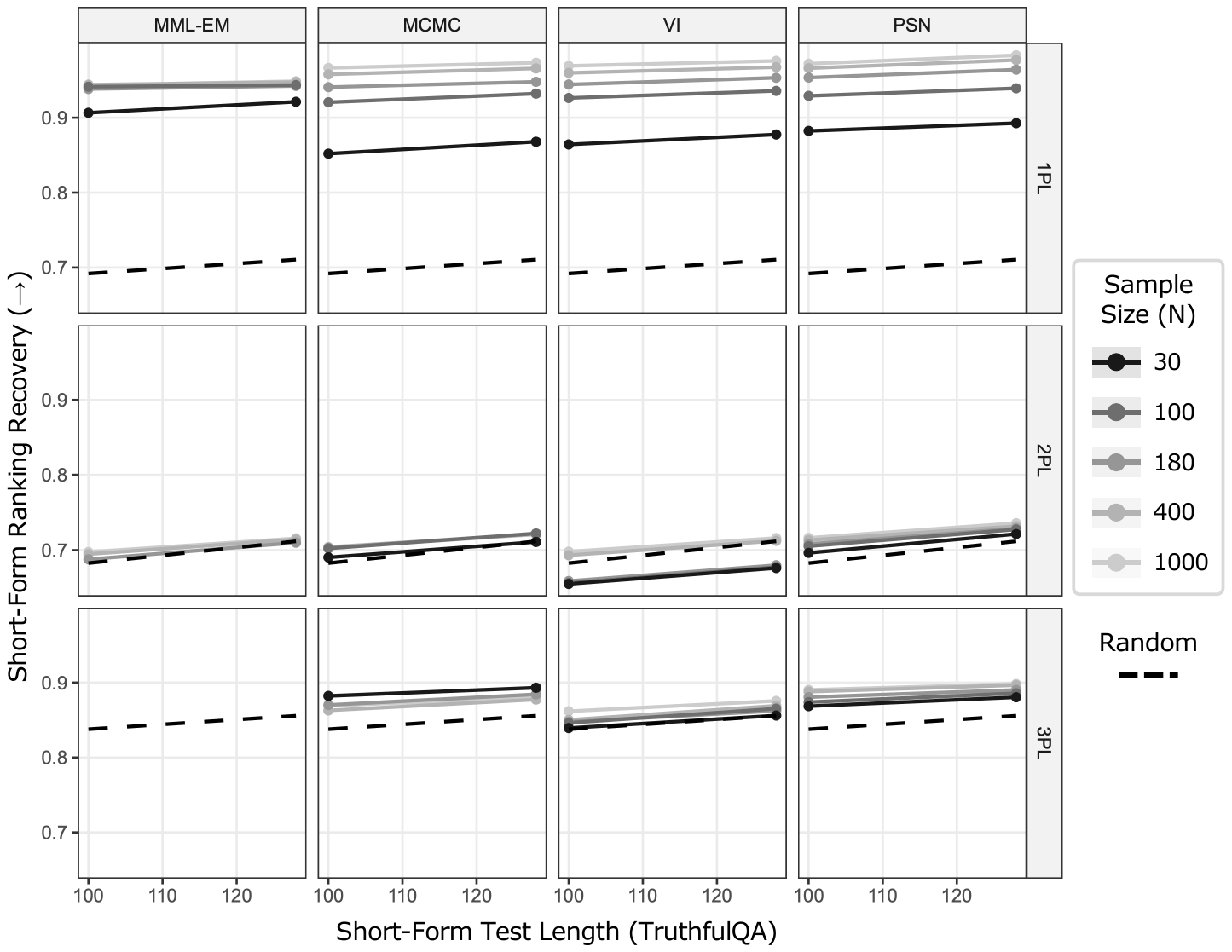}
    \caption{Short-form ranking recovery for larger subsets of \textsc{TruthfulQA} across IRT models, estimators, and sample sizes.}
    \label{fig:app_short_truthfulqa}
\end{figure*}

\begin{figure*}[ht]
    \centering
    \includegraphics[width=0.9\linewidth]{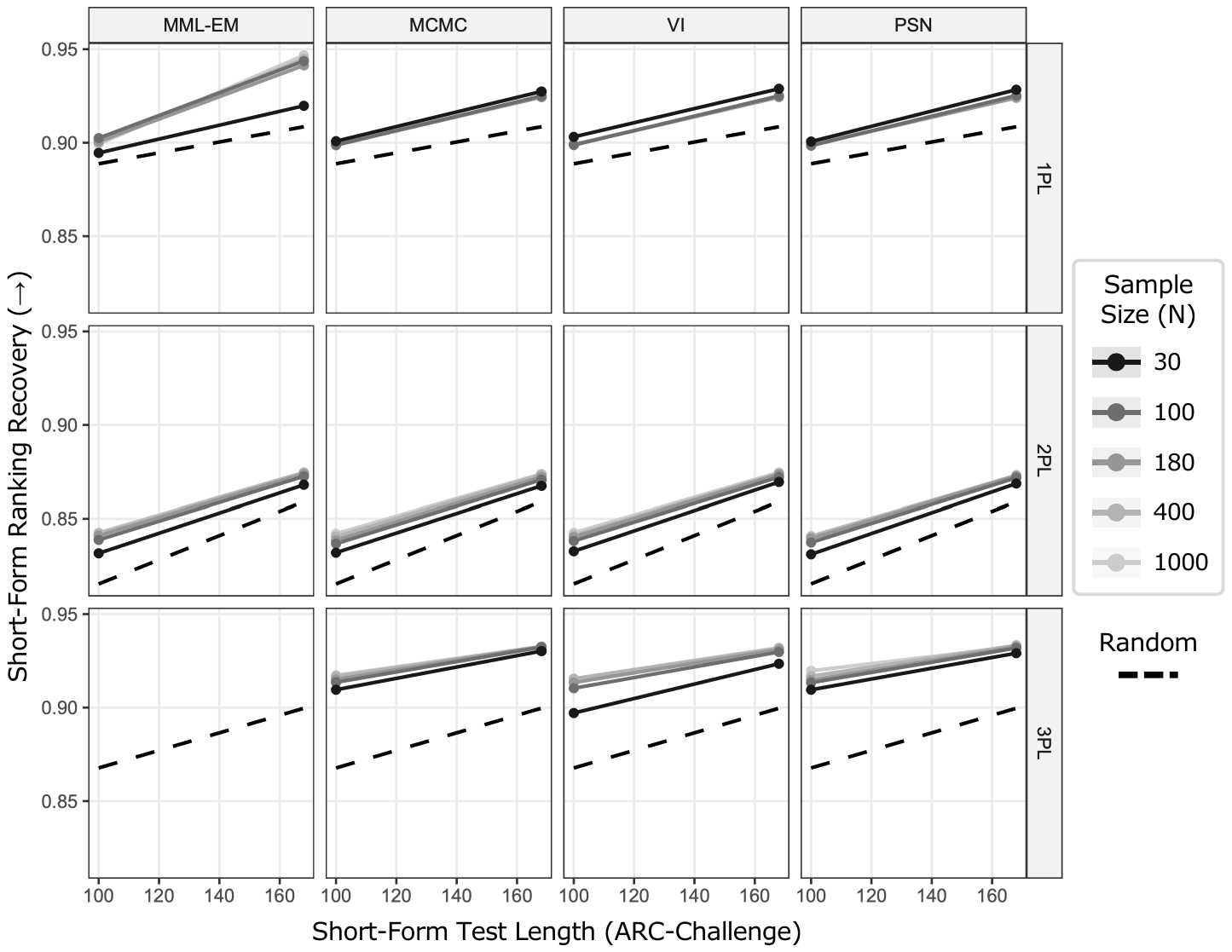}
    \caption{Short-form ranking recovery for larger subsets of \textsc{ARC-Challenge} across IRT models, estimators, and sample sizes.}
    \label{fig:app_short_arc}
\end{figure*}

\begin{figure*}[ht]
    \centering
    \includegraphics[width=0.9\linewidth]{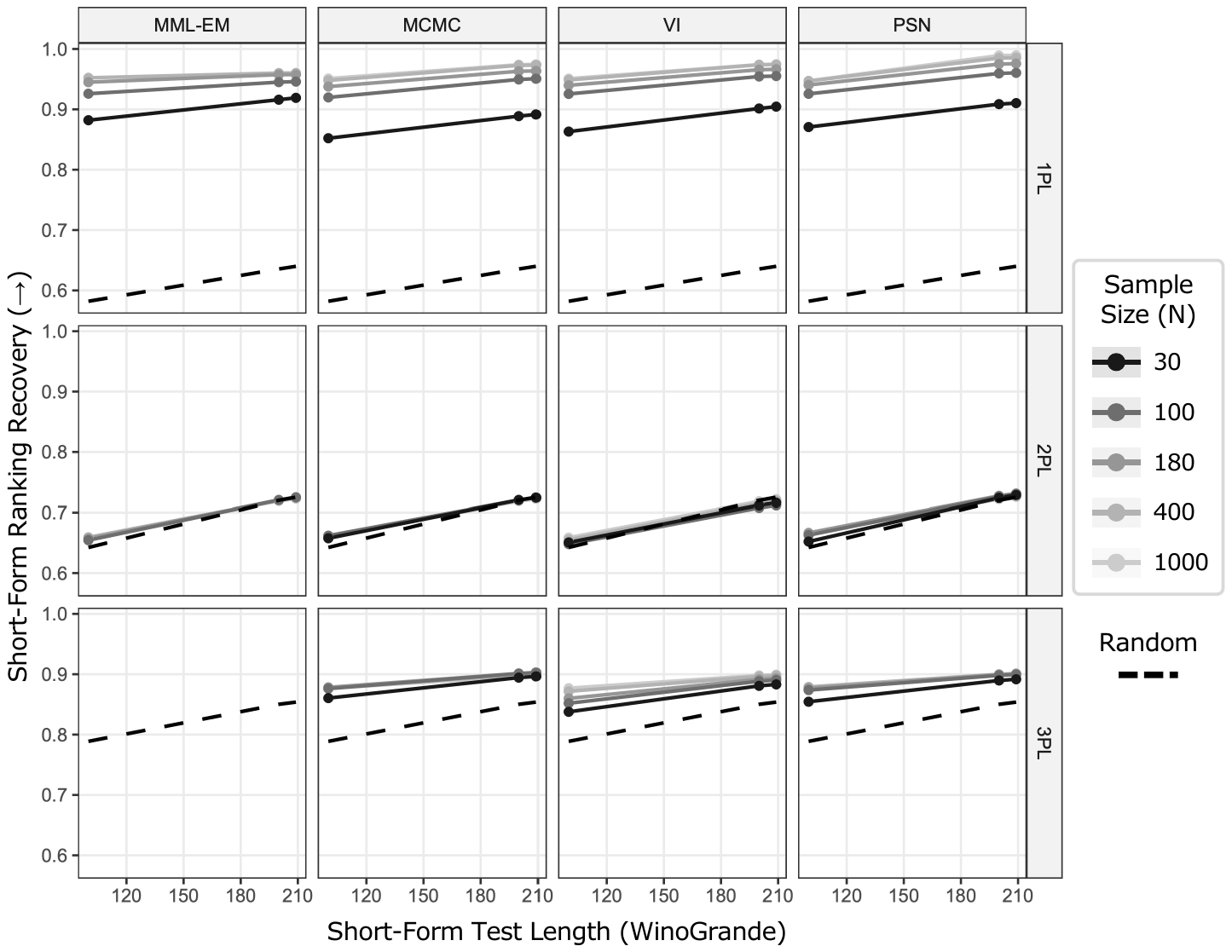}
    \caption{Short-form ranking recovery for larger subsets of \textsc{WinoGrande} across IRT models, estimators, and sample sizes.}
    \label{fig:app_short_winogrande}
\end{figure*}

\begin{figure*}[ht]
    \centering
    \includegraphics[width=0.9\linewidth]{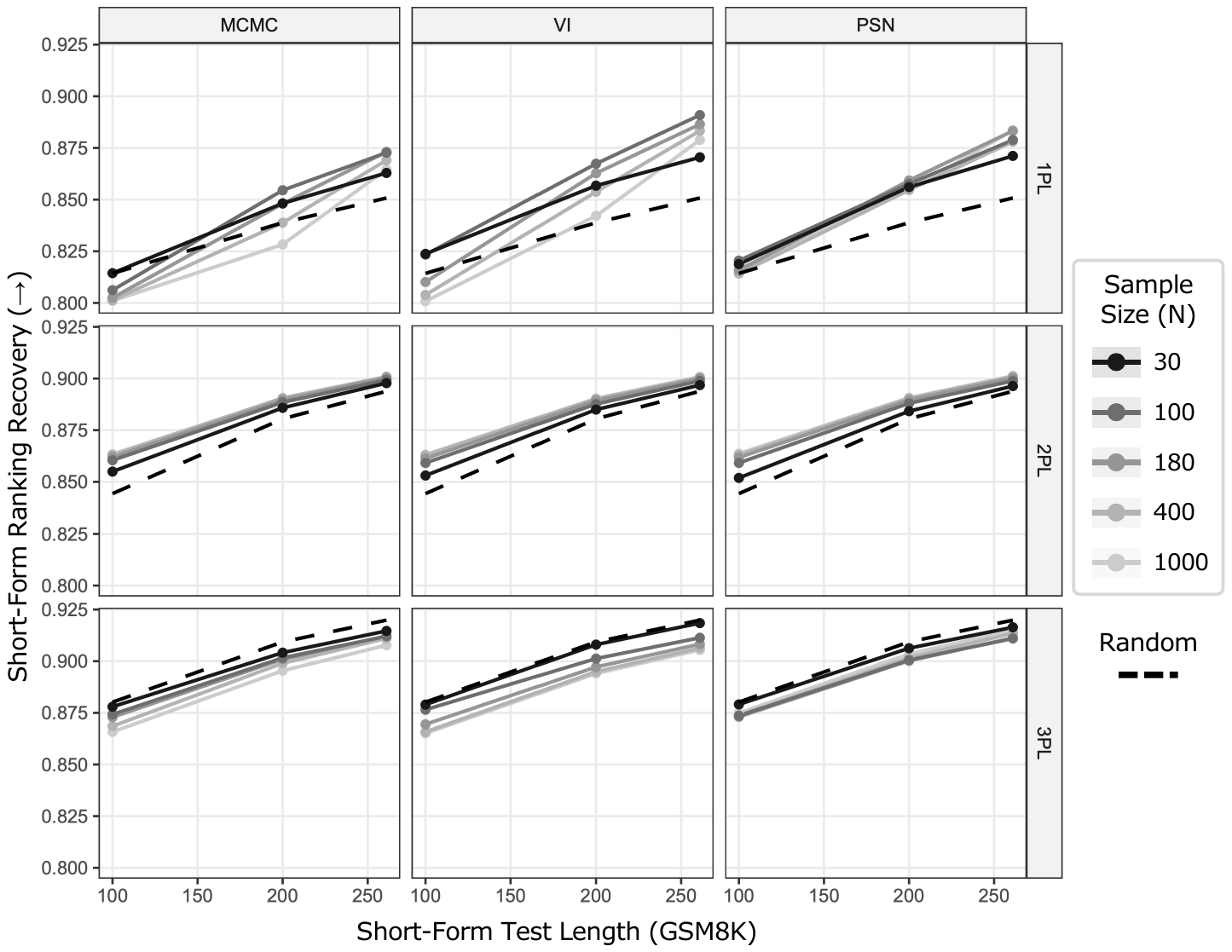}
    \caption{Short-form ranking recovery for larger subsets of \textsc{GSM8K} across IRT models, estimators, and sample sizes.}
    \label{fig:app_short_gsm8k}
\end{figure*}

\begin{figure*}[ht]
    \centering
    \includegraphics[width=0.9\linewidth]{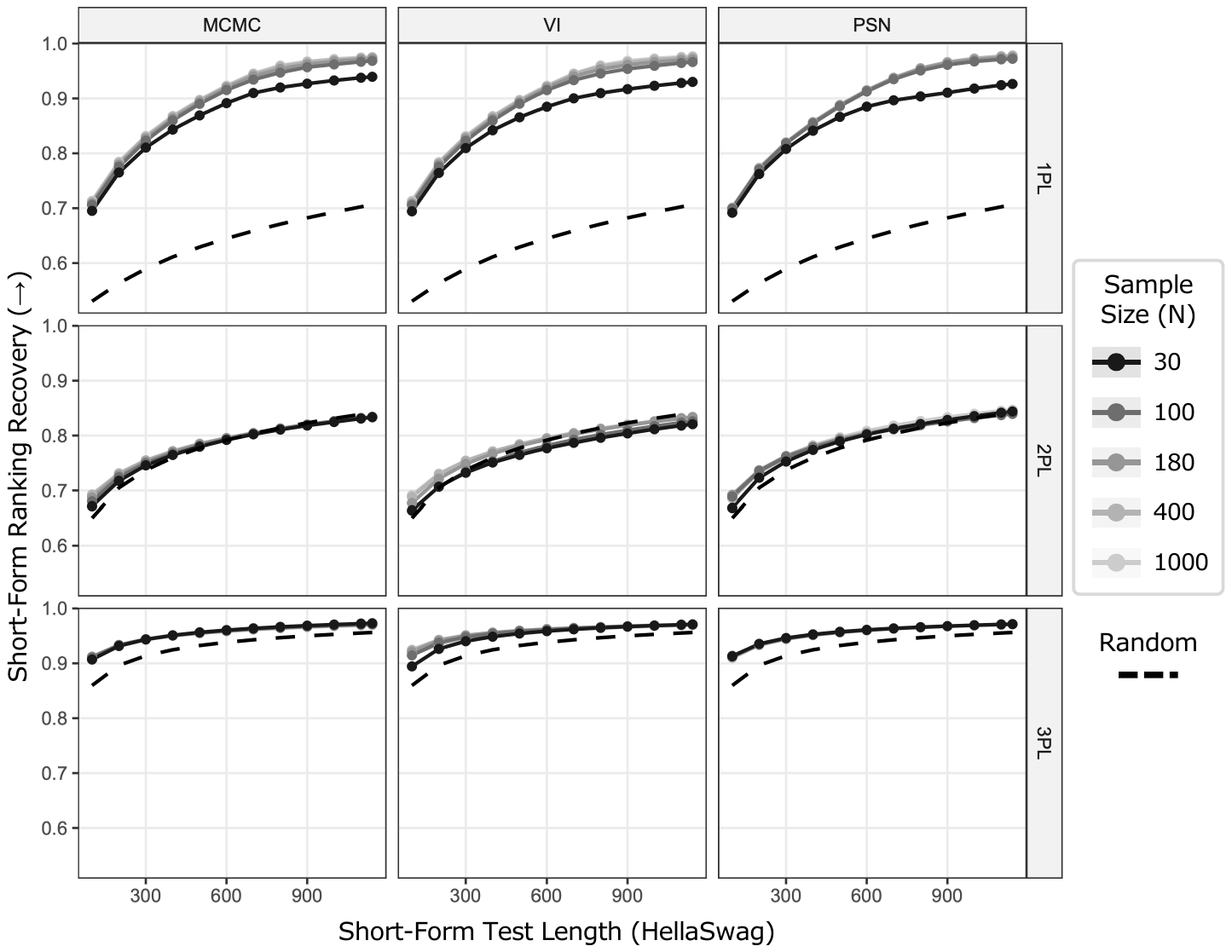}
    \caption{Short-form ranking recovery for larger subsets of \textsc{HellaSwag} across IRT models, estimators, and sample sizes.}
    \label{fig:app_short_hellaswag}
\end{figure*}

\begin{figure*}[ht]
    \centering
    \includegraphics[width=0.9\linewidth]{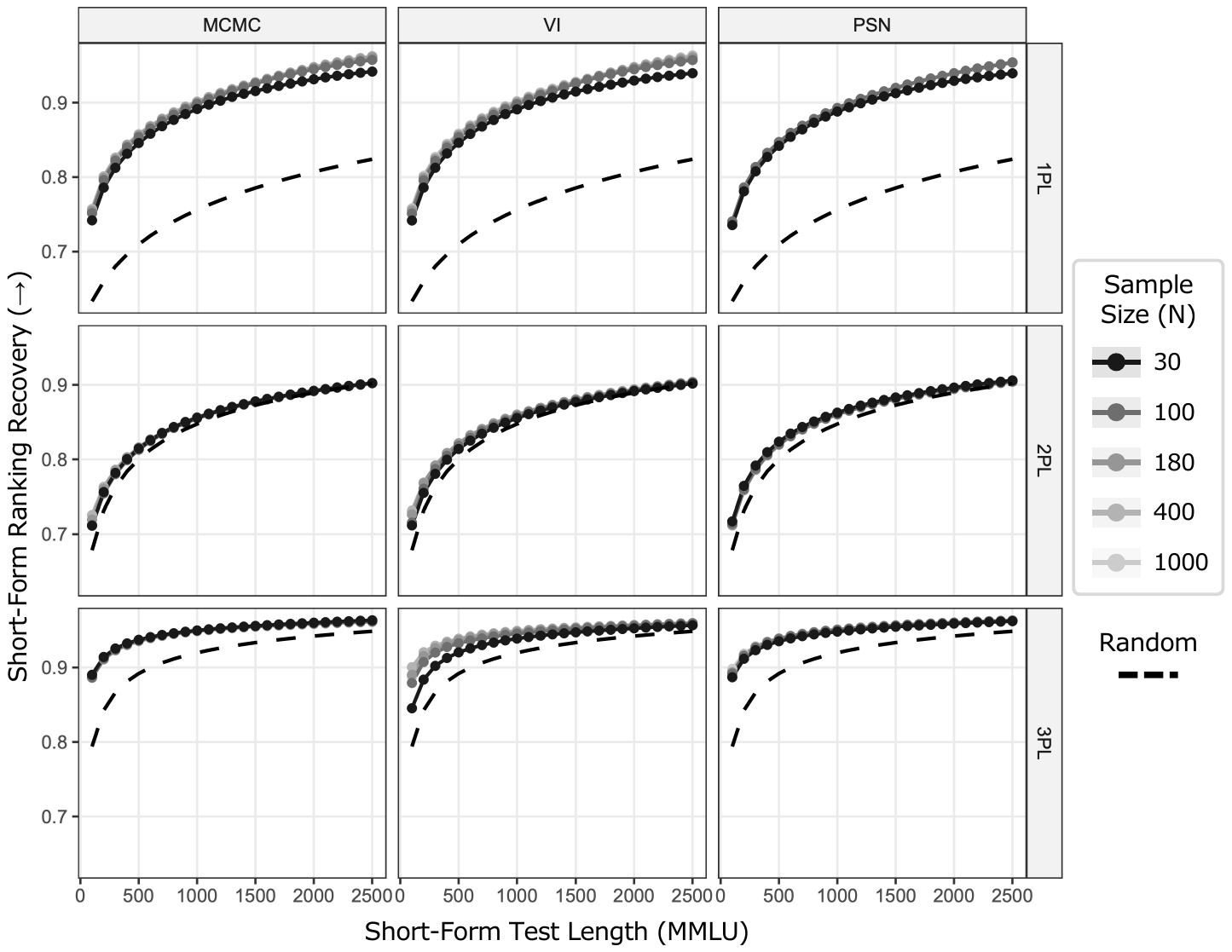}
    \caption{Short-form ranking recovery for larger subsets of \textsc{MMLU} across IRT models, estimators, and sample sizes.}
    \label{fig:app_short_mmlu}
\end{figure*}

Please refer to Fig.~\ref{fig:app_efficient_grid} for the complete short-form ranking-recovery results across the three test lengths (i.e., 10\%, 20\%, and 100 items), IRT models, benchmarks, and estimators. 
Figs.~\ref{fig:app_20p}~\&~\ref{fig:app_100} further break down the results by sample size.
Additional results across a broader range of short-form benchmark sizes are provided in Figs.~\ref{fig:app_short_truthfulqa}-\ref{fig:app_short_mmlu}.

\end{document}